\documentclass{article}

\PassOptionsToPackage{numbers, compress}{natbib}

 \usepackage[main, final]{neurips_2026}

\usepackage[utf8]{inputenc} 
\usepackage[T1]{fontenc}    
\usepackage{hyperref}       
\usepackage{url}            
\usepackage{graphicx}       
\usepackage{booktabs}       
\usepackage{multirow}       
\usepackage{amsmath}        
\usepackage{amsfonts}       
\usepackage{nicefrac}       
\usepackage{microtype}      
\usepackage{xcolor}         
\usepackage{makecell}
\usepackage{subcaption}

\title{Smooth Piecewise Cutting for Neural Operator to Handle Discontinuities and Sharp Transitions}

\author{
  Ha Dang\textsuperscript{a}, \ \ \
  Sebastian Schmidt\textsuperscript{a}, \ \ \
  Jürgen Hesser\textsuperscript{a,b,c,d,e} 
  \\[1ex]
  \textsuperscript{a}Mannheim Institute for Intelligent Systems in Medicine (MIISM), Heidelberg University \\
  \textsuperscript{b}Interdisciplinary Center for Scientific Computing (IWR), Heidelberg University \\
  \textsuperscript{c}Heidelberg Institute for Theoretical Studies (HITS), Heidelberg University \\
  \textsuperscript{d}Central Institute for Computer Engineering (ZITI), Heidelberg University \\
  \textsuperscript{e}CZS Heidelberg Initiative for Model-Based AI (MBAI), Heidelberg University \\
  Heidelberg, 69117, Germany \\[3pt]
  \texttt{\{ha.dang,Juergen.Hesser\}@medma.uni-heidelberg.de} \\
  \texttt{sebastian.schmidt@stud.uni-heidelberg.de}
}

\begin{document}

\maketitle

\label{sec:abstract}
\begin{abstract}
Neural operators have achieved strong performance in learning solution operators of partial differential equations (PDEs), but their inherently continuous representations struggle to capture discontinuities and sharp transitions. Existing approaches typically approximate such features within continuous function spaces, often requiring increased model capacity and high-resolution data. In this work, we propose Cut-DeepONet, a two-stage training framework that explicitly models discontinuities while reducing learning complexity. Our approach reformulates the problem via a lifting strategy, partitioning the domain into smooth subregions while representing discontinuities as boundaries in a higher-dimensional space. This separation aligns the operator learning task with the inductive bias of neural networks and avoids directly approximating discontinuities. An additional network predicts input-dependent discontinuity locations for unseen inputs, which are then used to guide the neural operator in generating smooth components within each region. Experiments on benchmark PDEs show that Cut-DeepONet outperforms state-of-the-art methods, even when trained on low-resolution datasets. The method excels on problems with discontinuities and sharp transitions, while using fewer trainable parameters. Our results highlight the benefits of changing the representation of operator learning rather than increasing model complexity.\footnote{Code available at \href{https://github.com/DangHa/cut-deeponet}{https://github.com/DangHa/cut-deeponet}}
\end{abstract}

\label{sec:introduction}
\section{Introduction}

Neural operators have recently emerged as a powerful framework for learning solution operators of differential equations, offering a promising complement to classical numerical solvers. Models such as DeepONet \cite{Lu2021deeponet} and the Fourier Neural Operator (FNO) ~\cite{li2020fourier} demonstrate strong empirical performance and are supported by universal approximation theorem for operator learning.

However, a fundamental limitation remains: standard neural operators struggle to accurately represent solutions with discontinuities or sharp transitions ~\cite{CHAUHAN2025100005}. Due to their underlying architectures, these models typically operate in smooth function spaces such as Sobolev spaces or reproducing kernel Hilbert spaces. As a result, they exhibit an inherent smoothing bias, which degrades performance when the ground-truth solution has large Sobolev norms—for example, in the presence of shock waves or discontinuities. While this smoothness bias is often desirable for regular problems, it significantly limits expressivity in many physically relevant regimes.

This limitation is well recognized in the literature. Lanthaler and Hadorn et al.~\cite{Lanthaler2023Nonlinear, Hadorn2022ShiftDeepONet} showed that the linear reconstruction in DeepONet makes it difficult to learn sharp transitions, particularly when the network responsible for learning basis functions has limited capacity. To address these challenges, a range of architectural improvements has been proposed. Nonlinear reconstruction strategies, such as those explored in~\cite{Lanthaler2023Nonlinear, Hadorn2022ShiftDeepONet}, as well as variants including NOMAD~\cite{seidman2022nomad}, flexDeepONet~\cite{venturi2023svddeeponet}, and HyperDeepONet~\cite{lee2024hyperdeeponet}, increase the expressivity of neural operators by enabling input-dependent basis functions. While these approaches alleviate some limitations of linear reconstruction, they still operate within continuous representations and therefore struggle to accurately capture discontinuities. Modeling discontinuities often requires specialized constructions or representations that are not naturally supported within standard neural operator frameworks; in particular, the network must be informed about the characteristics of these discontinuities. 

In this work, we take a different approach inspired by Lifting the Winding Number by Chang et al. \cite{Chang2025Lifting}, which introduces a lifting strategy to represent thin-wall discontinuities in neural fields. Rather than increasing model capacity or modifying architectures within the same representation, we change the representation in which the operator is learned. We reformulate the problem with discontinuities by lifting it into a higher-dimensional space, in which the domain is decomposed into multiple subdomains such that discontinuities occur only at their boundaries. A neural operator is then trained to learn the smooth solution components that satisfy the boundary conditions across adjacent subdomains. In this transformed representation, the output function becomes more regular, allowing the neural network to focus on learning smooth pieces within each subdomain while avoiding the need to model discontinuities.

For unseen inputs, the discontinuity locations or the sharp transitions are not known and not always be inferred analytically from the input data. Instead, they emerge from the nonlinear dynamics of the PDE and must be determined through the solution process, particularly in complex systems such as the Euler equations or in models exhibiting sharp transitions, such as the bidomain model \cite{HorgmoJaegeretal2023}. To address this issue, we introduce a discontinuity prediction network, called Cutting Net. This network partitions the domain via lifting into a higher-dimensional space, enabling the neural operator to generalize across varying discontinuity structures.

Based on these ideas, we propose Cut-DeepONet, a two-stage training framework (Figure ~\ref{fig:method-overview}). In the first stage, a network is trained to predict discontinuity locations. In the second stage, a neural operator is trained in the lifted domain to learn the smooth solution components within each subdomain. In prediction stage, the neural operator then generates these smooth pieces guided by the discontinuity locations predicted by the Cutting Net.

Our main highlights:
\begin{itemize}
    \item \textbf{Operator learning in smooth space}: Based on the lifting method in neural fields \cite{Chang2025Lifting}, we extend it to operator learning with input-dependent discontinuities. We introduce an additional network to generalize to unseen inputs and generalize the framework to capture sharp transitions.

    \item \textbf{Truly generate discontinuities in operator learning}: Our method is the first to explicitly generate dynamic  discontinuous solutions in operator learning, whereas existing approaches typically approximate such functions using high-resolution continuous representations.

    \item \textbf{Reduce operator learning complexity}: By ignoring the influence of discontinuities in operator learning task and focusing on smooth features aligned with the bias of neural networks, our method requires fewer trainable parameters and lower-resolution training data. Moreover, the training dataset can be generated using low-resolution numerical solvers, our approach can extrapolate near discontinuity boundaries without smoothing the discontinuity.

    \item \textbf{Outperformance on discontinuous operator learning tasks}: We demonstrate that the proposed method achieves higher accuracy and faster convergence than state-of-the-art approaches on problems involving discontinuities or sharp transitions (see Table ~\ref{tb:benchmark_table}).
\end{itemize}

\label{sec:related_work}
\section{Related Works}

In numerical methods, discontinuities remain challenging, as most schemes construct continuous approximations from discrete representations, often leading to instabilities or Gibbs oscillations near shocks \cite{Gottlieb1997gibbs}. To address this issue, high-resolution shock-capturing schemes, such as Essentially Non-Oscillatory (ENO) \cite{HARTEN19973} and Weighted Essentially Non-Oscillatory (WENO) \cite{LIU1994200} methods, have been developed to suppress non-physical oscillations while maintaining sharp resolution of discontinuities, albeit at increased computational cost. More recently, Beck et al.~\cite{BECK2020109824} proposed a neural-network-based shock detector for discontinuous Galerkin methods, using a convolutional neural network (CNN) to identify troubled regions and guide shock-capturing strategies.

In physics-informed neural networks (PINNs), where the solution is learned as a mapping from the domain to the solution space, discontinuities are often associated with domain separation. In such cases, each subdomain separated by a discontinuity can be learned using independent networks, as in XPINN \cite{kopani2024xpinn} and cPINN \cite{JAGTAP2020113028}. Alternatively, Hu et al. \cite{HU2022111576} and Zhao et al. \cite{ZHAO2025118184} demonstrated that discontinuities can be represented via a higher-dimensional embedding by introducing an additional dimension. Recently, Roy et al. \cite{roy2026phideeponetdiscontinuitycapturingneural} extended this idea to operator learning through the $\phi$-DeepONet. However, these approaches typically assume that the domain separation or discontinuity locations are known and fixed. In contrast, for operator learning of nonlinear PDEs, the locations and number of discontinuities are generally unknown. Furthermore, residual-based adaptive methods in PINNs \cite{Toscano2026residualmethod}, such as adaptive point weighting \cite{ANAGNOSTOPOULOS2024116805} and adaptive collocation point refinement \cite{JIAO2024108770} in high-error regions, are widely used to improve accuracy near discontinuities. These approaches can also be extended to DeepONet by introducing adaptive weighting and sampling strategies across different regions of the solution domain, thereby improving the representation of high-frequency features, as demonstrated in R-adaptive DeepONet \cite{zhu2024radaptivedeeponetlearningsolution}.

In operator learning, beyond architectures such as Shift-DeepONet, NOMAD, flexDeepONet, and HyperDeepONet, alternative designs have been proposed to better capture discontinuities and high-frequency features. For instance, the Wavelet Neural Operator \cite{TRIPURA2023115783} leverages wavelet transforms, while the more recent Walsh–Hadamard Neural Operator \cite{cavallazzi2025walshhadamardneuraloperatorssolving} employs Walsh–Hadamard transforms, both introducing alternative spectral bases that are better suited for representing high-frequency regions and discontinuities. Roseberry et al. \cite{OLEARYROSEBERRY2024112555} and Qiu et al. \cite{qiu2024derivative} incorporate derivative information as auxiliary data. When high-resolution solutions are available, including derivative information in the loss function can further improve approximation accuracy, particularly near discontinuities. Chawla et al. \cite{chawla2026discontinuousgalerkinfiniteelement} introduced a hybrid framework, the Discontinuous Galerkin Finite Element Operator Network (DG-FEONet), which integrates neural operator learning with discontinuous Galerkin discretizations, enabling physics-informed, data-free training for problems with discontinuous coefficients. Roohi et al. \cite{roohi2025shockawarephysicsguidedfusiondeeponetoperator} extended the Fusion-DeepONet \cite{peyvan2025fusiondeeponetdataefficientneuraloperator} by incorporating physics-informed loss terms to enhance learning in shock regions. Furthermore, Kolmogorov–Arnold Networks (KANs) \cite{liu2024kankolmogorovarnoldnetworks} and their operator-learning variant, DeepOKAN \cite{ABUEIDDA2025117699}, recently provide a new paradigm for function and operator approximation. localized edge functions can represent rapid oscillations more directly than fixed activation functions, potentially improving the approximation of high-frequency behavior.

\label{sec:problem}
\section{Neural Operator \& Limitation for Discontinuous Problems}

\subsection{Neural Operator}

Operator learning aims to approximate mappings between infinite-dimensional function spaces. Let $\{u_i(x)\}_{i=1}^N$ denote a set of input functions with $x \in \Omega_x$, and let the corresponding outputs be $\{G(u_i)(y)\}_{i=1}^N$ with $y \in \Omega_y$. The operator $G: \mathcal{U} \rightarrow \mathcal{V}$ maps a function $u(x)$ to another function $G(u)(y)$. The learning task is therefore: given paired samples $\{(u_i, G(u_i))\}_{i=1}^N$, learn an approximation $\hat{G}$ such that
\begin{equation}
    \hat{G}(u)(y) \approx G(u)(y), \quad \forall u \in \mathcal{U}, \; y \in \Omega_y.
\end{equation}
Neural operators parameterize this approximation $\hat{G}$ using neural networks, with the key property of generalizing across different discretizations of both the input function $u$ and the output locations $y$:
\begin{equation}
    \hat{G}_\theta: u(x) \mapsto \hat{G}_\theta(u)(y).
\end{equation}
Lu et al.~\cite{Lu2021deeponet} introduced the DeepONet architecture, which represents the operator using two subnetworks: a branch network $b(\cdot)$ and a trunk network $t(\cdot)$. Given sampled evaluations of the input function $u$, the operator is approximated as
\begin{equation}
    \hat{G}_\theta(u)(y) = b(u(x)) \cdot t(y),
\end{equation}
where $b(u(x)) \in \mathbb{R}^p$ encodes the input function and $t(y) \in \mathbb{R}^p$ encodes the output domain. The final prediction is obtained via the dot product of these representations. In contrast, Li et al.~\cite{li2020fourier} introduced FNO, which parameterizes the operator in the spectral domain. The model iteratively updates a function representation, with each layer given by
\begin{equation}
    v_{k+1}(x) = \sigma\left( W v_k(x) + \mathcal{F}^{-1} \left( R \cdot \mathcal{F}(v_k)(\xi) \right)(x) \right),
\end{equation}
where $\mathcal{F}$ and $\mathcal{F}^{-1}$ denote the Fourier and inverse Fourier transforms, $R$ is a learned spectral kernel, and $W$ is a pointwise linear transformation. After $L$ layers, the final representation provides the operator approximation $\hat{G}_\theta(u)(y) \approx v_L(y)$.

DeepONet allows flexible sampling via its two subnetworks, whereas the FNO typically relies on a fixed discretization grid for both input and output. However, the final linear combination in DeepONet can be a limitation, as it may struggle to accurately represent discontinuities and sharp transitions, as demonstrated in the work of Lanthaler and Hadorn et al.~\cite{Lanthaler2023Nonlinear, Hadorn2022ShiftDeepONet}.

\subsection{Limitation: A Continuous Representation for Discontinuous Functions}

Lu et al.~\cite{Lu2021deeponet} and Kovachki et al.~\cite{fno_universal} respectively established universal approximation theorems for DeepONet and FNO. In particular, suppose the output space is the Banach space of continuous functions $\mathcal{V} = C(\Omega_y),$
equipped with the supremum norm $\|f\|_{C(\Omega_y)}=\max_{y\in\Omega_y} |f(y)|$. For a continuous nonlinear operator $G:\mathcal{U}\to C(\Omega_y)$ and any $\varepsilon>0$, there exists a parameterized neural operator $\hat{G}_{\theta}$ such that
\begin{equation}
    \| G(u)(y) - \hat{G}_\theta(u)(y) \| < \varepsilon, \quad \forall u \in K \subset \mathcal{U}, \; \forall y \in \Omega_y,
\end{equation}
where $K$ is a compact subset of $\mathcal{U}$. However, if $G(u)(y)$ is discontinuous as a function of $y$, then $G(u)(y)\notin C(\Omega_y)$. Therefore, the theorem does not directly apply when $G(u)(y)$ has discontinuities.

Gottlieb et al. \cite{Gottlieb1997gibbs} showed that the Gibbs phenomenon occurs when expansion methods, such as Fourier expansions, are used to approximate functions with discontinuities. Reducing this effect typically requires increasing the number of high-frequency basis functions. A similar issue arises in neural operators, which represent mappings as continuous functions and therefore approximate discontinuities as sharp but still smooth transitions. Accurately capturing these transitions is challenging, as it requires larger model capacity and high-resolution training data to resolve the steep gradients near discontinuities. The problem becomes harder in operator learning, where the location and shape of discontinuities vary across inputs and domains. In such cases, the model requires a significantly larger basis, as discussed by Lanthaler et al. \cite{Lanthaler2023Nonlinear}.

Even though discontinuities only appear in small regions, they require a lot of effort to learn and can reduce accuracy in nearby smooth areas. This shows a key limitation: current neural operators cannot truly represent discontinuities. Therefore, we need new methods that can directly represent discontinuities in solutions, reduce the effort needed to learn them, and at the same time keep high accuracy in the smooth parts of the solution.

\label{sec:method}
\section{Proposed Method}

\begin{figure}[ht]
    \centering
    \includegraphics[width=\linewidth]{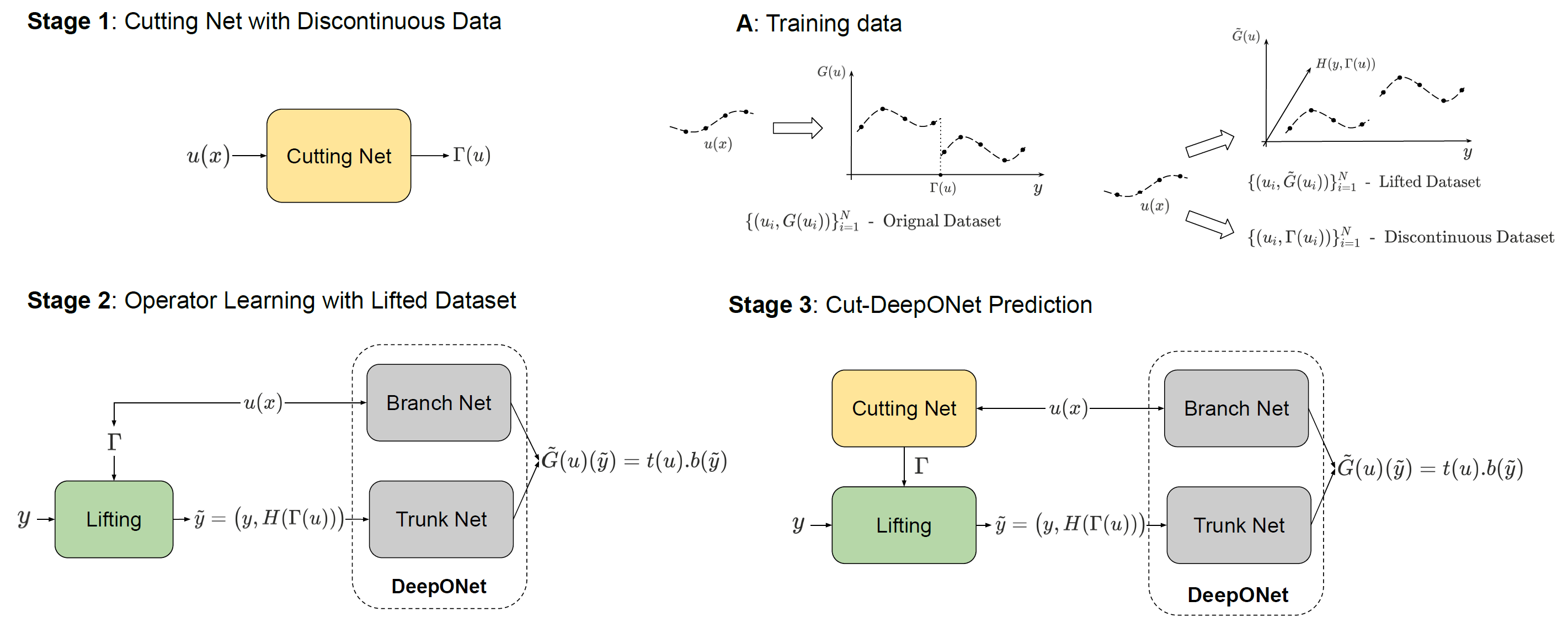}
    \caption{\textbf{Overview of the Cut-DeepONet architecture and training procedure}. (\textbf{A}) From the original dataset, the discontinuity locations are extracted from $G(u)(y)$ and used to construct both a Lifted Dataset and a Discontinuous Dataset. \textbf{Stage 1}: The Cutting Net is trained on the Discontinuous Dataset to predict the locations of discontinuities. \textbf{Stage 2}: The Lifted Dataset is used to guide the operator learning model to generate smooth solution components. \textbf{At prediction}, the trained networks from both stages are combined to form Cut-DeepONet, which is able to generate discontinuous solutions for unseen inputs.}
    \label{fig:method-overview}
\end{figure}
To address these limitations, we view the discontinuous problem as the combination of several smooth pieces. Instead of forcing the neural operator to learn the discontinuities directly, we design the learning process so that the network ignores the discontinuities and focuses on learning the smooth parts. To do so, we propose an approach with two steps: predicting the discontinuity locations and learning the solution within each smooth subdomain. A network is introduced to estimate discontinuity locations, and the problem is then lifted to a higher-dimensional space where the solution becomes continuous. In this lifted space, the operator learning model only needs to approximate smooth functions, which can later be mapped back to recover the original discontinuous solution (Figure \ref{fig:method-overview}). As a result, discontinuities are naturally reconstructed at the boundaries between pieces.

\subsection{Lifting Method} 

The lifting approach in Chang et al. (Appendix \ref{sc:lifting_chang}) was designed for computer graphics problems with a smooth interface and fixed discontinuity locations. In operator learning, however, the setting is more general: the discontinuity set may evolve over time, depend on the input, and the solution may exhibit different behavior on each side of the discontinuity. Thus, we extend the lifting approach to operator learning problems with discontinuities as follows.

Let $\Gamma(u) \subset \Omega_y$ denote the discontinuity set associated with an input $u$. The goal of the lifting is to transform a discontinuous function $G(u)(y)$ into a function that is continuous in an augmented space. We define an additional coordinate
\begin{equation*}
H : \Gamma(u) \rightarrow \mathbb{R}^k,
\end{equation*}
which encodes the geometry of the discontinuity. Based on this, we introduce the augmented coordinate
\begin{equation}
\tilde{y} = \bigl(y, H(\Gamma(u))\bigr) \in \Omega_y \times \mathbb{R}^k.
\end{equation}
The lifted function is then defined as
\begin{equation}
\tilde{G}(u)(\tilde{y}) := G(u)(y).
\end{equation}
By construction, $\tilde{G}(u)$ is continuous with respect to $\tilde{y}$, even though $G(u)$ may be discontinuous in $y$.

If the discontinuity is represented by an open curve and the regions on both sides belong to the same underlying continuous surface, using the generalized winding number to define $H(\Gamma(u))$ is reasonable.

In operator learning, however, the discontinuity locations may form several closed regions, and the solution may exhibit different behavior in the regions separated by these regions. If different regions are assigned the same augmented value, they may be artificially connected in the lifted space, which is undesirable when the solution is discontinuous across those regions. Therefore, distinct labels are needed to distinguish the different regions. Especially for sharp transitions, we define two cutting locations corresponding to the beginning and end of each high-derivative region, and the solution behavior before and after the cut is completely different. If these regions are assigned the same value of $H(\Gamma(u))$, the lifting fails to hide the sharp jumps from the operator learning task. Therefore, in this case, instead of using the winding number as an additional coordinate, we assign distinct labels to each region. Let $\{\Omega_j\}_{j=1}^M$ denote the connected regions of $\Omega_y \setminus \Gamma(u)$. Then
\begin{equation}
H(\Gamma(u))(y) = c_j, \quad \text{for } y \in \Omega_j,
\end{equation}
where $c_i \neq c_j$ for $i \neq j$, ensuring that each region is uniquely represented in the augmented space.

\subsection{Data Preparation: Discontinuity Extraction}
In neural operator learning, the dataset is given as $\{(u_i, G(u_i))\}_{i=1}^N$. For both the lifting and prediction stages, the discontinuity locations $\Gamma(u_i)$ are required. These locations can be obtained analytically when the problem is sufficiently simple; otherwise, they can be extracted from the solution $G(u_i)$ in the original dataset. Based on this, two new datasets are constructed: a lifted dataset $\{(u_i, \tilde{G}(u_i))\}_{i=1}^N$, where $\tilde{G}(u_i)(\tilde{y}) = G(u_i)(y)$ with $\tilde{y} = (y, H(\Gamma(u_i)))$, and a discontinuous dataset $\{(u_i, \Gamma(u_i))\}_{i=1}^N$.

Analytically, for certain nonlinear first-order hyperbolic PDEs in conservation form, the method of characteristics can determine when and where shocks form through the intersection of characteristic curves. After shock formation, the Rankine–Hugoniot condition provides the shock speed by enforcing conservation across the discontinuity, as illustrated for the inviscid Burgers’ equation in  Appendix \ref{sc:analytically}. However, this framework does not generalize to arbitrary ODEs or PDEs.

In cases where such analytical tools are not available, discontinuities can instead be extracted directly from the solution, provided high-resolution data from numerical simulations or exact solutions are available. This can be done by identifying peaks in the derivative field. However, when only low-resolution numerical solutions are available, the discontinuity is often smeared into a continuous transition. In this situation, it is necessary not only to estimate the discontinuity location, but also to remove the artificial smoothing introduced by the numerical solver, so that the model does not learn an incorrect continuous representation. For sharp transitions, the start and end points of the sharp region can be identified by locating the maximum derivative and then extending in both directions until the solution returns to a smooth region. We demonstrate several examples in Appendix \ref{sc:extract_discontinuity}.

\subsection{Cut-DeepONet}
\textbf{Stage 1: Training the Cutting Net:} \ \
Before applying the lifting step, it is necessary to identify the discontinuity locations for unseen inputs. To this end, we learn a mapping from the input space to the discontinuity set using the discontinuous dataset $\{(u_i, \Gamma(u_i))\}_{i=1}^N$. More precisely, we define the Cutting Net as
\begin{equation}
\text{cnet}_\theta : \mathcal{U} \rightarrow \mathcal{D}, \quad
\text{cnet}_\theta(u) := \Gamma(u),
\end{equation}
for any $u \in \mathcal{U}$, where $\mathcal{D}$ denotes the space of discontinuity locations, and $\text{cnet}_\theta$ denotes the Cutting Net.

\textbf{Stage 2: Operator Learning in Smooth Higher Dimensional Space:} \ \
After lifting, the original discontinuous operator $G$ is transformed into a continuous operator $\tilde{G}$ defined on the augmented space. Using the lifted dataset $\{(u_i, \tilde{G}(u_i))\}_{i=1}^N$, the goal is to learn the mapping
\begin{equation}
\tilde{G} : \mathcal{U} \rightarrow C(\Omega \times \mathbb{R}^k),
\end{equation}
where $\tilde{G}(u)(\tilde{y})$ is continuous with respect to $\tilde{y} = (y, H(\Gamma(u))(y))$.
The universal approximation theorem for operators now applies in the lifted space. We approximate this operator using a neural operator $\hat{G}_\phi$, defined as
\begin{equation}
\hat{G}_\phi : \mathcal{U} \rightarrow C(\Omega \times \mathbb{R}^k), \quad
\hat{G}_\phi(u)(\tilde{y}) \approx \tilde{G}(u)(\tilde{y}),
\end{equation}
where $\phi$ denotes the learnable parameters. Since $\tilde{G}(u)$ is smooth in the lifted space, $\hat{G}_\phi$ can be trained more efficiently without needing to approximate discontinuities.

\textbf{Cut-DeepONet Prediction:} \ \
The trained components from Stage 1 and Stage 2 are combined to form the Cut-DeepONet. To get the solution in the original domain, simply evaluate the lifted representation at the corresponding augmented coordinate.
\begin{equation}
G(u)(y) = \hat{G}_\phi\bigl(u\bigr)\bigl(y, H(cnet_\theta(u))\bigr).
\end{equation}
Cut-DeepONet generates a solution with a true discontinuity at the predicted discontinuity location.

\label{sec:experiment}
\newpage
\section{Experiments}
In this section, we consider three benchmark problems exhibiting discontinuities and sharp transitions to demonstrate how such features can be extracted from data. We compare our proposed method against representative approaches designed to handle these challenges. We evaluate all methods using standard metrics and provide a detailed analysis of the results.

\subsection{Problems}
\textbf{Linear Advection:} \ \
We consider the linear advection equation with two discontinuities, similar to the setup in Lanthaler et al.~\cite{Lanthaler2023Nonlinear}. However, we employ a lower spatial resolution while extending the range of cases in the time domain. The dataset is generated on the domain $x \in [0,1]$ with spatial resolution $N_x = 500$, and $t \in [0, 0.25]$ with time resolution $N_t = 15$. We further investigate training under even lower resolutions (Figure~\ref{fig:example_linear_advection}). A total of 500 samples with varying initial conditions are generated, of which 90\% are used for training. The variability is introduced through the wave height, starting location, and width, covering approximately $20\%$ of the possible parameter space.

\textbf{Inviscid Burgers' Equation:} \ \
We consider the inviscid Burgers' equation with a single discontinuity in the initial condition, leading to shock formation. The dataset is generated on the domain $x \in [-1,1]$ with spatial resolution $N_x = 1000$, and $t \in [0, 0.5]$ with temporal resolution $N_t = 1000$. A total of 250 samples is generated ($\approx 20\%$ cases), each defined by a piecewise constant initial condition with a single discontinuity. The variability is introduced through the left state $u_L \in [1, 2]$ and the initial discontinuity location $x_d \in [-0.9, -0.1]$, while the right state is fixed as $u_R = 0$.

\textbf{Parsimonious Model:} \ \
This is a reduced Hodgkin--Huxley model for rabbit ventricular cardiomyocytes, characterized by a sharp transition in the opening of Na$^+$ channels following the stimulus. We generate 200 samples with varying stimulus parameters for training. The stimulus is designed to induce a full action potential within a time window of 400~ms by varying the onset time, amplitude, and duration of the applied stimulus. $N_t = 80000$ is data resolution, but we just use 2000 point sampling training. Further details of those equations and data generation process are provided in Appendix~\ref{sc:problem_appendix}.

\subsection{Compared Methods}
The proposed Cut-DeepONet is compared with DeepONet, FlexDeepONet, HyperDeepONet, and FNO (Appendix~\ref{sc:different_methods}), as these are among the most widely used methods for handling discontinuities in neural operators. All models are evaluated under the architectural settings summarized in Tables~\ref{tb:setting_table_1}, \ref{tb:setting_table_2}, and \ref{tb:setting_table_3}, corresponding to the linear advection equation, inviscid Burgers' equation, and the parsimonious model, respectively.

\subsection{Metrics}
In this work, the $L_1$ loss is used to compare different methods. In addition, a $\textit{Dis}$ loss is introduced, defined as the $L_1$ loss evaluated within a localized region around the discontinuity. This metric measures how the presence of a discontinuity affects the accuracy in nearby smooth regions for different methods.
\begin{align}
\mathcal{L}_{L_1} &= \frac{1}{|\Omega|} \int_{\Omega} \left| G(u)(y) - \hat{G}_\phi(u)(y) \right| \, dy, \\
\mathcal{L}_{\textit{Dis}} &= \frac{1}{|\Omega_{\textit{Dis}}|} \int_{\Omega_{\textit{Dis}}} \left| G(u)(y) - \hat{G}_\phi(u)(y) \right| \, dy.
\end{align}
where $\Omega_{\textit{Dis}}$ denotes the region near the discontinuity. In this paper, 10\% of the domain is used to define this region.

A discontinuity-aware metric such as $\mathcal{L}_{\textit{Dis}}$ more closely indicates the error in approximating discontinuities and their effects on the surrounding regions. However, both $L_1$ and $\textit{Dis}$ losses can underestimate the performance of the proposed method. When Cut-DeepONet generates a true discontinuity parallel to the ground truth, the error is evaluated on both sides of the jump, effectively doubling the contribution. In contrast, continuous methods smooth over the discontinuity and intersect it, resulting in a smaller measured error. This leads to an overestimation of the error for the proposed method, especially when the discontinuity gap is large. This issue is exacerbated when using the $L_2$ loss; therefore, we do not employ it. A detailed discussion is provided in Appendix~\ref{sc:metric_problem}. To address this in future evaluations, one potential direction is to adjust the metric for Cut-DeepONet by halving the accumulated $L_1$ error between the predicted and ground-truth cut locations to remove the double-counting effect.
\subsection{Results}

\begin{figure}[ht]
    \centering
    \begin{subfigure}{\textwidth}
        \centering
        \includegraphics[width=\textwidth]{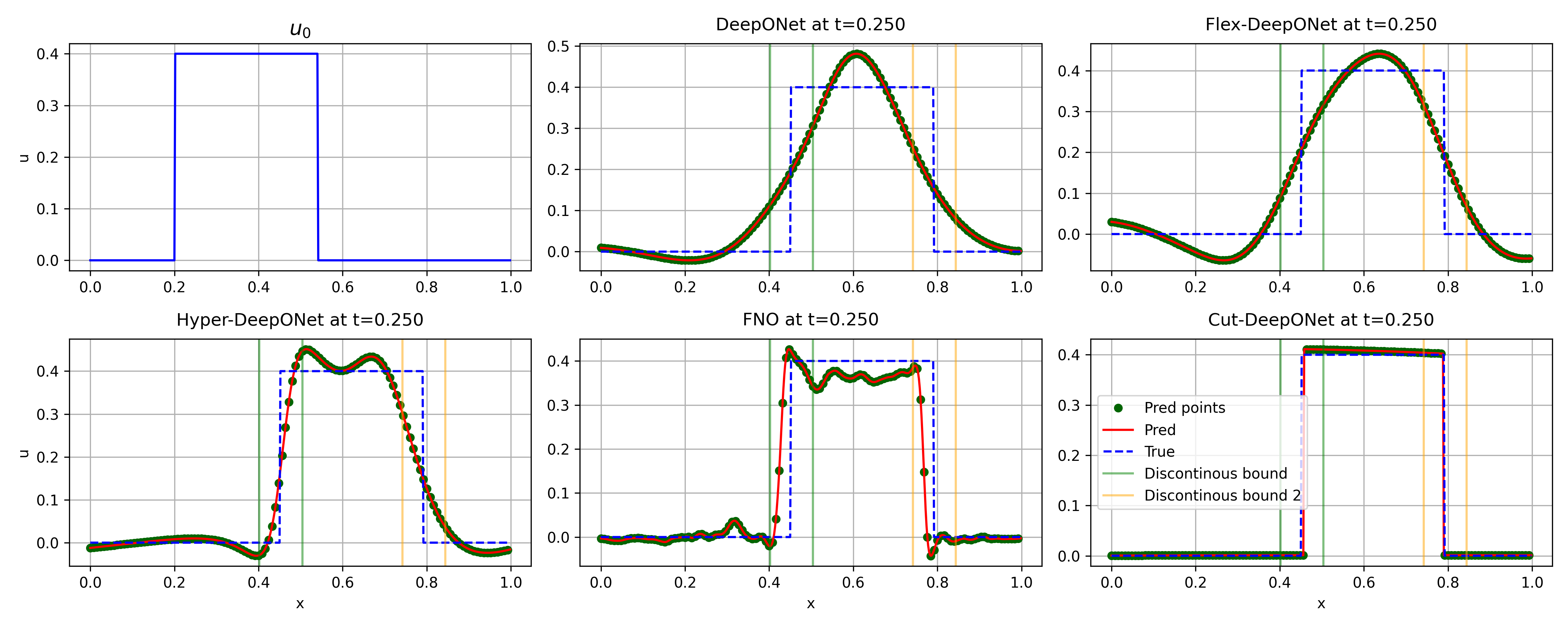}
        \caption{The performance of different methods on a linear advection example. Cut-DeepONet generates piecewise solutions with true discontinuities. Training is performed using data with a resolution of 2000.}
        \label{fig:linear_advection_compare}
    \end{subfigure}
    \begin{subfigure}{0.495\textwidth}
        \centering
        \includegraphics[width=\textwidth]{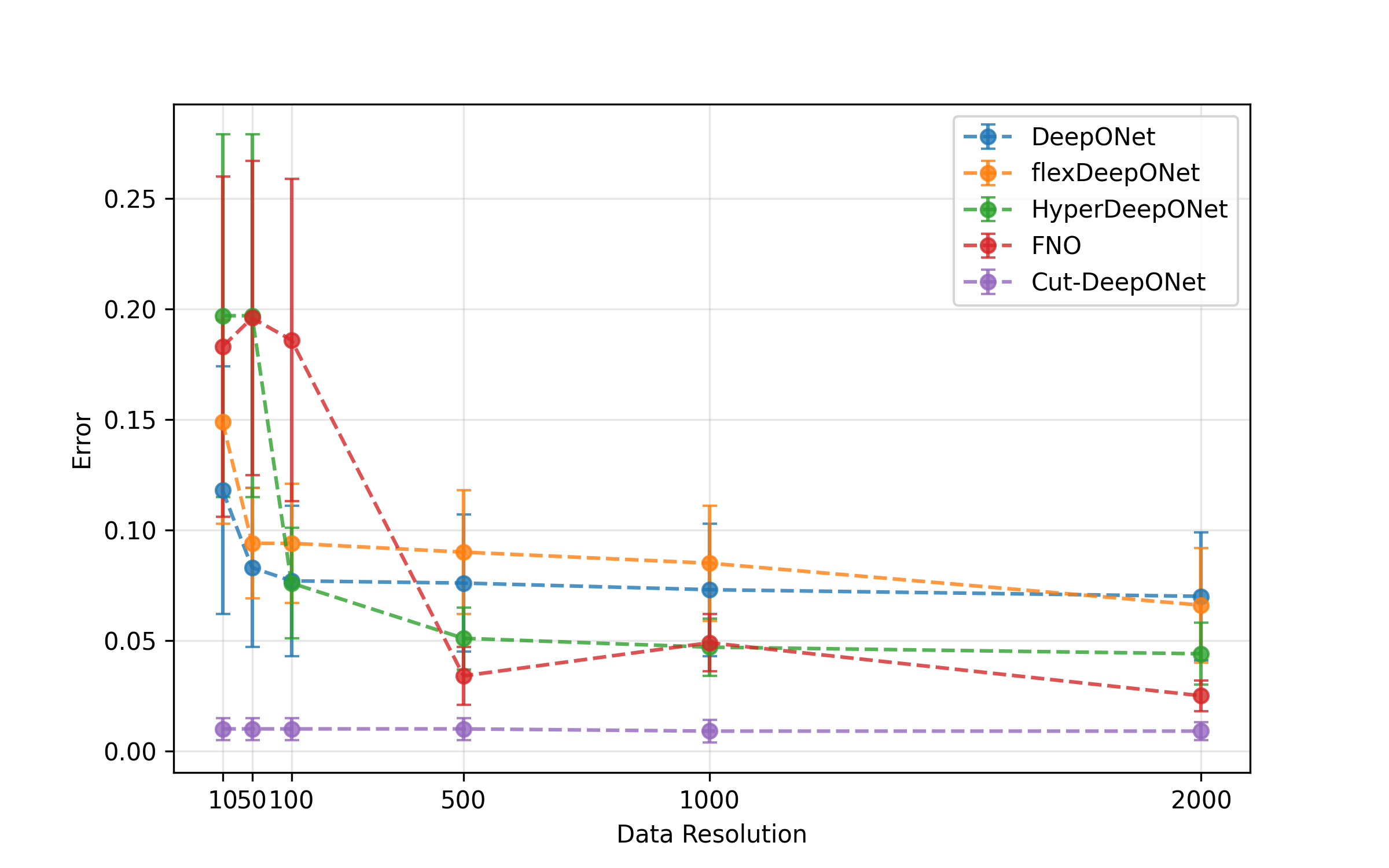}
        \caption{Error for different resolutions}
        \label{fig:l1_linear_advection}
    \end{subfigure}
    \begin{subfigure}{0.495\textwidth}
        \centering
        \includegraphics[width=\textwidth]{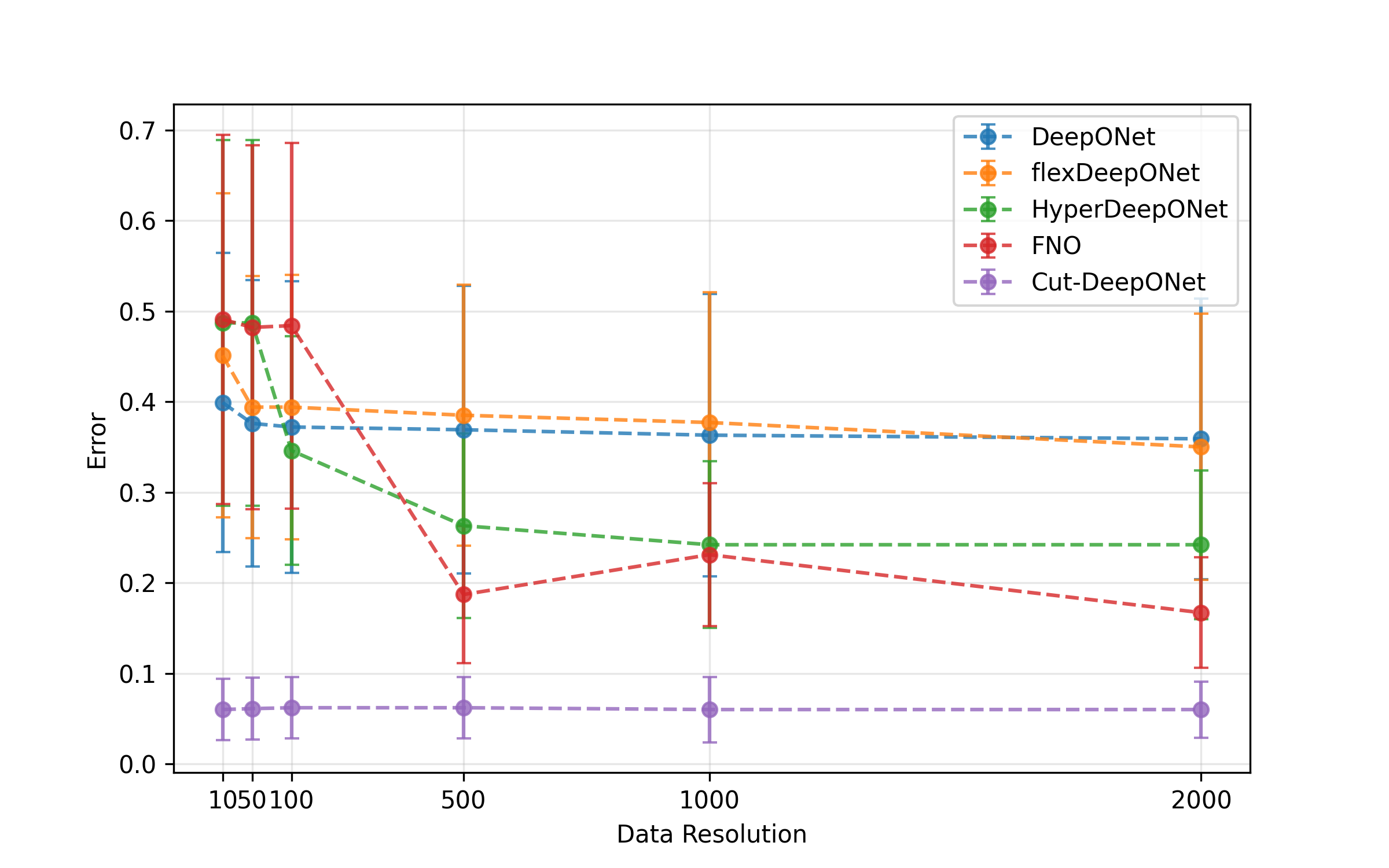}
        \caption{Error near discontinuities for different resolutions}
        \label{fig:l1_dis_linear_advection}
    \end{subfigure}

    \caption{Cut-DeepONet outperforms other methods on the linear advection equation and maintains strong performance even with low-resolution training data.}
    \label{fig:example_linear_advection}
\end{figure}

\begin{table}[h]
\centering
\small
\setlength{\tabcolsep}{4.2pt}
\begin{tabular}{llcccccc}
\toprule
&  & \thead{\textbf{DeepONet}} & \thead{\textbf{Flex-}\\\textbf{DeepONet}} & \thead{\textbf{Hyper-}\\\textbf{DeepONet}} & \thead{\textbf{FNO}} & \thead{\textbf{Cut-}\\\textbf{DeepONet} (our)} \\
\midrule \midrule

\multirow{2}{*}{\textbf{\shortstack[l]{Linear \\ Advection}}} 
    & $L_1$        & 0.070 $\pm$ 0.029 & 0.066 $\pm$ 0.026 &  0.044 $\pm$ 0.014 & 0.025 $\pm$ 0.007 & \textbf{0.009 $\pm$ 0.004}  \\
    & \textit{Dis}       & 0.359 $\pm$ 0.155 & 0.350 $\pm$ 0.147 & 0.242 $\pm$ 0.082 & 0.167 $\pm$ 0.061  & \textbf{0.060 $\pm$ 0.031} \\ 
\midrule

\multirow{2}{*}{\textbf{\shortstack[l]{Inviscid \\ Burgers'}}} 
    & $L_1$        & 0.089 $\pm$ 0.026 & 0.062 $\pm$ 0.027 & 0.031 $\pm$ 0.006 & 0.060 $\pm$ 0.035 & \textbf{0.010 $\pm$ 0.007}  \\
    & \textit{Dis}       & 0.407 $\pm$ 0.100 & 0.193 $\pm$ 0.035  & 0.135 $\pm$ 0.031 & 0.279 $\pm$ 0.110  & \textbf{0.053 $\pm$ 0.037} \\ 
\midrule


\multirow{2}{*}{\shortstack[l]{\textbf{Parsimonious} \\ \textit{(sharp jump)}}} 
    & $L_1$        & 6.116 $\pm$ 1.900 & 9.382 $\pm$ 4.283 & 6.744 $\pm$ 3.095 & 6.287 $\pm$ 7.049 & \textbf{2.032 $\pm$ 0.830}  \\
    & \textit{Dis}       & 19.592 $\pm$ 3.335 & 15.077 $\pm$ 2.322  & 5.669 $\pm$ 2.082 & \textbf{1.777 $\pm$ 0.856}  & 2.050 $\pm$ 0.588 \\ 
\bottomrule
\end{tabular}
\caption{\textbf{Cut-DeepONet outperforms other methods on problems with discontinuities and sharp transitions}. All results are evaluated on a separate high-resolution test set. In the parsimonious setting, it achieves better overall performance, although the discontinuity error (\textit{Dis}) is higher than FNO due to under-sampling of region for sharp transition (Figure~\ref{fig:pasimonious_result}). Additionally, the $L_1$ loss can underestimate the performance of Cut-DeepONet (Appendix \ref{sc:metric_problem}).}
\label{tb:benchmark_table}
\end{table}

\begin{table}[ht]
\centering
\scriptsize
\setlength{\tabcolsep}{2.5pt}
\renewcommand{\arraystretch}{1.12}

\resizebox{\linewidth}{!}{%
\begin{tabular}{llcccccc}
\toprule
& &
\multirow{2}{*}{\textbf{Ground truth}}
& \multicolumn{5}{c}{\textbf{Discontinuous Dataset for Cutting Net}} \\
\cmidrule(lr){4-8}
& &
& \textbf{+ 0\% Noise}
& \textbf{+ 1\% Noise}
& \textbf{+ 5\% Noise}
& \textbf{+ 10\% Noise}
& \textbf{+ 20\% Noise} \\
\midrule
\midrule

\multirow{2}{*}{\textbf{\shortstack[l]{Linear \\ Advection}}}
& \makecell{{Cutting Net} \\ {($L_2$ \& \%)}}
& 0\%
& \makecell{$(0.04 \pm 0.08)e^{-3}$ \\ $0.02 \pm 0.03\%$}
& \makecell{$(0.06 \pm 0.08)e^{-3}$ \\ $0.02 \pm 0.03\%$}
& \makecell{$(0.08 \pm 0.14)e^{-3}$ \\ $0.03 \pm 0.05\%$}
& \makecell{$(0.26 \pm 0.59)e^{-3}$ \\ $0.09 \pm 0.21\%$}
& \makecell{$(1.26 \pm 4.11)e^{-3}$ \\ $0.44 \pm 1.45\%$}
\\

\cmidrule(lr){2-8}

& \makecell{{Cut-DeepONet} \\ {($L_1$ \& \textit{Dis})}}
& \makecell{$0.005 \pm 0.003$ \\ $0.021 \pm 0.010$}
& \makecell{$0.008 \pm 0.003$ \\ $0.051 \pm 0.016$}
& \makecell{$0.010 \pm 0.004$ \\ $0.064 \pm 0.037$}
& \makecell{$0.011 \pm 0.005$ \\ $0.076 \pm 0.042$}
& \makecell{$0.016 \pm 0.007$ \\ $0.124 \pm 0.063$}
& \makecell{$0.028 \pm 0.018$ \\ $0.216 \pm 0.117$}
\\

\midrule

\multirow{2}{*}{\textbf{\shortstack[l]{Inviscid \\ Burgers'}}}
& \makecell{{Cutting Net} \\ {($L_2$ \& \%)}}
& 0\%
& \makecell{$(0.05 \pm 0.07)e^{-3}$ \\ $0.04 \pm 0.05\%$}
& \makecell{$(0.07 \pm 0.18)e^{-3}$ \\ $0.05 \pm 0.09\%$}
& \makecell{$(0.05 \pm 0.09)e^{-3}$ \\ $0.04 \pm 0.06\%$}
& \makecell{$(0.14 \pm 0.27)e^{-3}$ \\ $0.09 \pm 0.19\%$}
& \makecell{$(0.15 \pm 0.32)e^{-3}$ \\ $0.11 \pm 0.22\%$}
\\

\cmidrule(lr){2-8}

& \makecell{{Cut-DeepONet} \\ {($L_1$ \& \textit{Dis})}}
& \makecell{$0.007 \pm 0.005$ \\ $0.022 \pm 0.007$}
& \makecell{$0.011 \pm 0.007$ \\ $0.058 \pm 0.025$}
& \makecell{$0.011 \pm 0.007$ \\ $0.065 \pm 0.043$}
& \makecell{$0.011 \pm 0.007$ \\ $0.057 \pm 0.028$}
& \makecell{$0.013 \pm 0.009$ \\ $0.083 \pm 0.059$}
& \makecell{$0.013 \pm 0.007$ \\ $0.085 \pm 0.035$}
\\

\midrule

\multirow{2}{*}{\textbf{Parsimonious}}
& \makecell{{Cutting Net} \\ {($L_2$ \& \%)}}
& 0\%
& \makecell{$(0.41 \pm 1.50)e^{3}$ \\ $0.00 \pm 0.00\%$}
& \makecell{$(1.54 \pm 3.50)e^{3}$ \\ $0.00 \pm 0.01\%$}
& \makecell{$(10.89 \pm 30.17)e^{3}$ \\ $0.03 \pm 0.07\%$}
& \makecell{$(27.34 \pm 69.03)e^{3}$ \\ $0.06 \pm 0.16\%$}
& \makecell{$(77.61 \pm 154.16)e^{3}$ \\ $0.18 \pm 0.36\%$}
\\

\cmidrule(lr){2-8}

& \makecell{{Cut-DeepONet} \\ {($L_1$ \& \textit{Dis})}}
& \makecell{$2.032 \pm 0.829$ \\ $2.051 \pm 0.595$}
& \makecell{$2.032 \pm 0.829$ \\ $2.053 \pm 0.583$}
& \makecell{$2.028 \pm 0.825$ \\ $2.015 \pm 0.579$}
& \makecell{$2.051 \pm 0.854$ \\ $2.235 \pm 0.788$}
& \makecell{$2.086 \pm 0.845$ \\ $2.544 \pm 1.252$}
& \makecell{$2.244 \pm 0.890$ \\ $4.115 \pm 2.414$}
\\

\bottomrule
\end{tabular}
}
\caption{
\textbf{Sensitivity analysis of Cutting Net and Cut-DeepONet errors under different noise levels effect in the Discontinuous Dataset.} The first column uses the ground-truth discontinuity locations without the Cutting Net. As noise increases, the Cutting Net error and final Cut-DeepONet error also increase. These experiments demonstrate the robustness and characteristic behavior of the Cut-DeepONet approach. Cut-DeepONet focuses on generating smooth solution segments and is therefore relatively insensitive to errors in the predicted cut locations. Although the discontinuity may shift to the left or right, the overall solution shape is preserved (Figure~\ref{fig:cutting_net}).
}
\label{tb:noise_robustness}

\end{table}

\textbf{Outperforming baselines on discontinuous problems with low-resolution data:} \ \
Table \ref{tb:benchmark_table} shows that Cut-DeepONet consistently achieves the best performance across both metrics on Linear Advection and Inviscid Burgers' equations. In particular, it significantly reduces both the global $L_1$ error and \textit{Dis} error, demonstrating its ability to accurately generate discontinuities while maintaining high accuracy in smooth regions. This is also illustrated in Figure \ref{fig:example_linear_advection}, where Cut-DeepONet is the only method that generates truly discontinuities in solution, while other methods exhibit oscillations or smoothing effects near the discontinuities. This demonstrates the effectiveness of our method and validates our hypothesis that learning smooth components is easier, and that the Cutting Net can guide the neural operator to generate discontinuities. Furthermore, Table~\ref{tb:noise_robustness} demonstrates the robustness of Cut-DeepONet in generating smooth solution segments, which are largely unaffected by errors in the predicted cut locations.

\textbf{Outperforming baselines on a sharp-transition problem:} \ \
In the parsimonious model with sharp transitions, Cut-DeepONet still achieves the lowest overall $L_1$ error. However, its \textit{Dis} error is higher than that of FNO. This is mainly due to the use of random sampling during training, where regions near discontinuities occupy only a small portion of the domain and are therefore underrepresented (Figure~\ref{fig:pasimonious_result}). As a result, these regions are not learned as effectively. A more balanced sampling strategy that ensures sufficient coverage near discontinuities could further improve performance in such settings.

\textbf{Learning from Low-Resolution Data with Few Trainable Parameters:} \ \
In addition, Tables~\ref{tb:setting_table_1}, \ref{tb:setting_table_2}, and \ref{tb:setting_table_3} report the number of trainable parameters for each model used in the benchmarks (Table~\ref{tb:benchmark_table}). Despite having fewer parameters, Cut-DeepONet still outperforms competing methods. Although the benchmark results are reported at relatively high resolution, Cut-DeepONet maintains strong performance even when the resolution is reduced. Notably, this holds despite the use of the $L_1$ loss, which is less favorable for discontinuous predictions.


\label{sec:conclusion}
\section{Conclusion and Discussion}

\textbf{Summary:} \
In this paper, we introduced Cut-DeepONet, a two-stage training framework for handling discontinuities and sharp transitions in neural operator learning. By separating the domain into smooth pieces and placing discontinuities at their boundaries in a higher-dimensional space, the proposed method is able to generate true discontinuities in the solution. This approach also reduces the complexity of operator learning by avoiding the need to approximate discontinuities directly. The framework is flexible and can be integrated with different variants of DeepONet, where the operator learning component in Stage 2 may be replaced by models such as HyperDeepONet or FlexDeepONet.

\textbf{Limitation:} \ \
There are several promising directions that could further improve the proposed approach. When separating regions with sharp transitions, the transition area usually occupies only a small portion of the domain, so adaptive sampling strategies or preprocessing methods could be introduced to ensure sufficient training points in every region. Extending the method to FNO also appears feasible, although additional modifications may be needed since the lifting process produces nonuniform-grid inputs that are not directly compatible with its standard formulation. In the current experiments, the Cutting Net is designed for a fixed number of discontinuities, but more flexible output representations, such as binary grids or mask predictions, could allow the method to handle varying numbers of discontinuities over time. Furthermore, the Cutting Net guides the Neural Operator in determining where discontinuities should be generated, allowing the Neural Operator to learn effectively even from low-resolution numerical data. However, the Cutting Net depends strongly on the accuracy of the training data; if the numerical solver provides incorrect discontinuity locations, the model will inherit these errors (Figure \ref{fig:cutting_net}). Therefore, it may be beneficial to train the Cutting Net with a physics-informed approach. Additionally, alternative loss functions for evaluating parallel discontinuous modeling should also be considered to enable fairer comparisons for discontinuous problems (Figure \ref{fig:metric_problem}). Moreover, extending the method to higher dimensions is feasible (Figure \ref{fig:2d_linear_advection}). The computational cost of Cut-DeepONet scales as $\mathcal{O}(n^{d+1})$, where $n$ is the number of sampling points and $d$ is the spatial dimension, with $n >> d$. However, Cut-DeepONet can operate effectively with a relatively small $n$. Further investigation of the performance of the Cutting Net is also needed for cases in which the discontinuity has a more complex geometry or a curved shape.

\section{Acknowledgments}
This research was supported by the Zentrales Innovationsprogramm Mittelstand (ZIM) under funding codes KK5044710AB3 and KK5085803AB3, and in part by the CZS Heidelberg Initiative for Model-Based AI under Grant P2021-02-001. We also acknowledge support from the data storage service SDS@hd, provided by the Ministry of Science, Research and the Arts Baden-Württemberg (MWK) and the German Research Foundation (DFG) through grants INST 35/1314-1 FUGG and INST 35/1503-1 FUGG. The authors gratefully acknowledge all sources of support.

The authors declare no potential conflicts of interest with respect to the research, authorship, and/or publication of this article.

\bibliography{liratures.bib}
\bibliographystyle{unsrtnat}

\label{sec:appendix}
\appendix

\newpage
\section{Compared method architectures}
\label{sc:different_methods}

\begin{figure}[h]
    \centering


    \begin{subfigure}{0.32\textwidth}
        \centering
        \includegraphics[width=\textwidth]{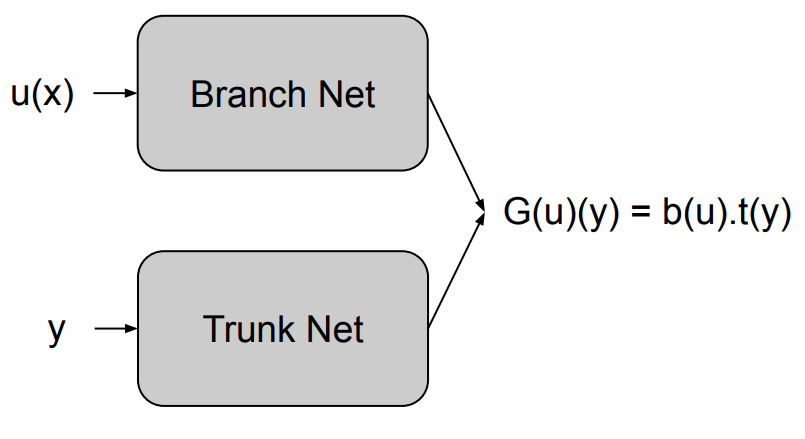}
        \caption{DeepONet \cite{Lu2021deeponet}}
        \label{fig:l1_linear_advection}
    \end{subfigure}
    \begin{subfigure}{0.32\textwidth}
        \centering
        \includegraphics[width=\textwidth]{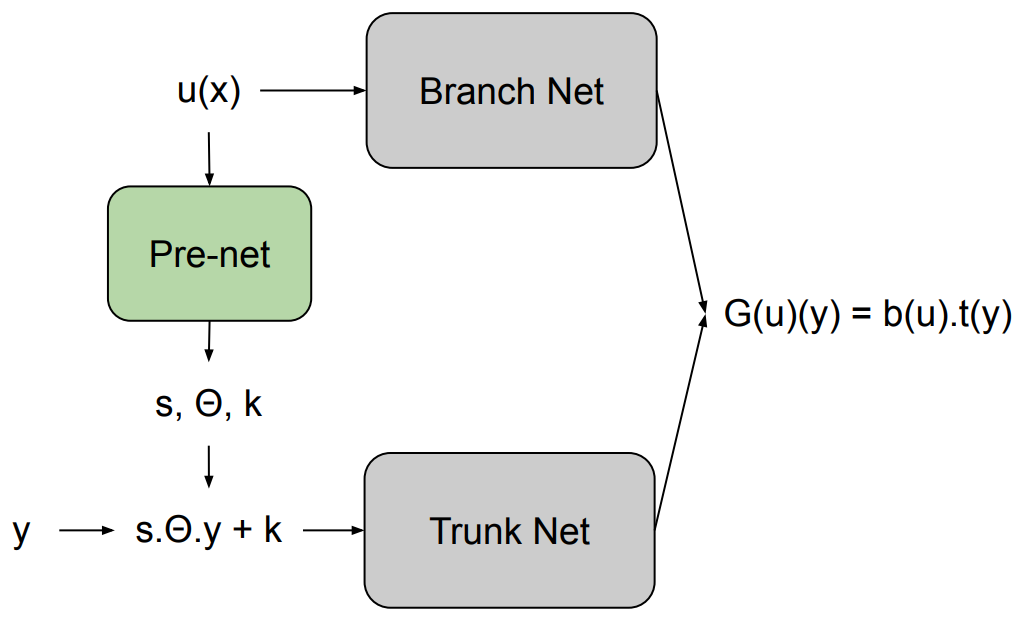}
        \caption{flexDeepONet \cite{venturi2023svddeeponet}}
        \label{fig:l1_dis_linear_advection}
    \end{subfigure}
    \begin{subfigure}{0.32\textwidth}
        \centering
        \includegraphics[width=\textwidth]{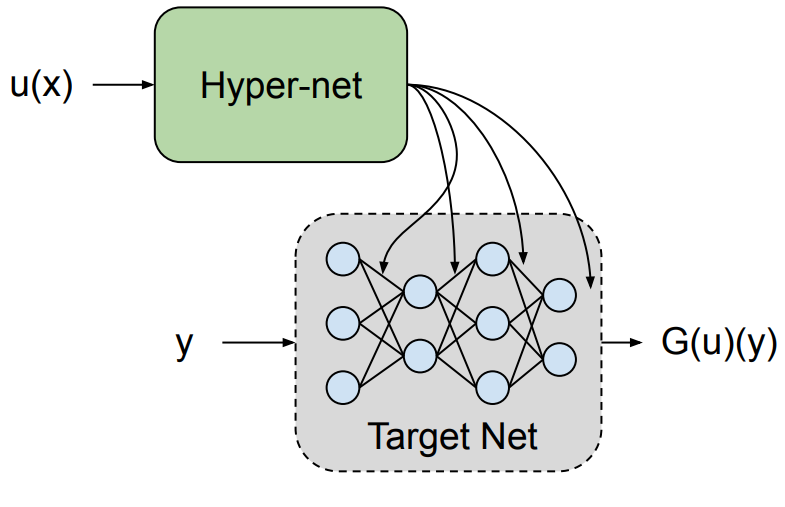}
        \caption{HyperDeepONet \cite{lee2024hyperdeeponet}}
        \label{fig:l1_dis_linear_advection}
    \end{subfigure}
    
    \medskip
    \begin{subfigure}{\textwidth}
        \centering
        \includegraphics[width=0.8\textwidth]{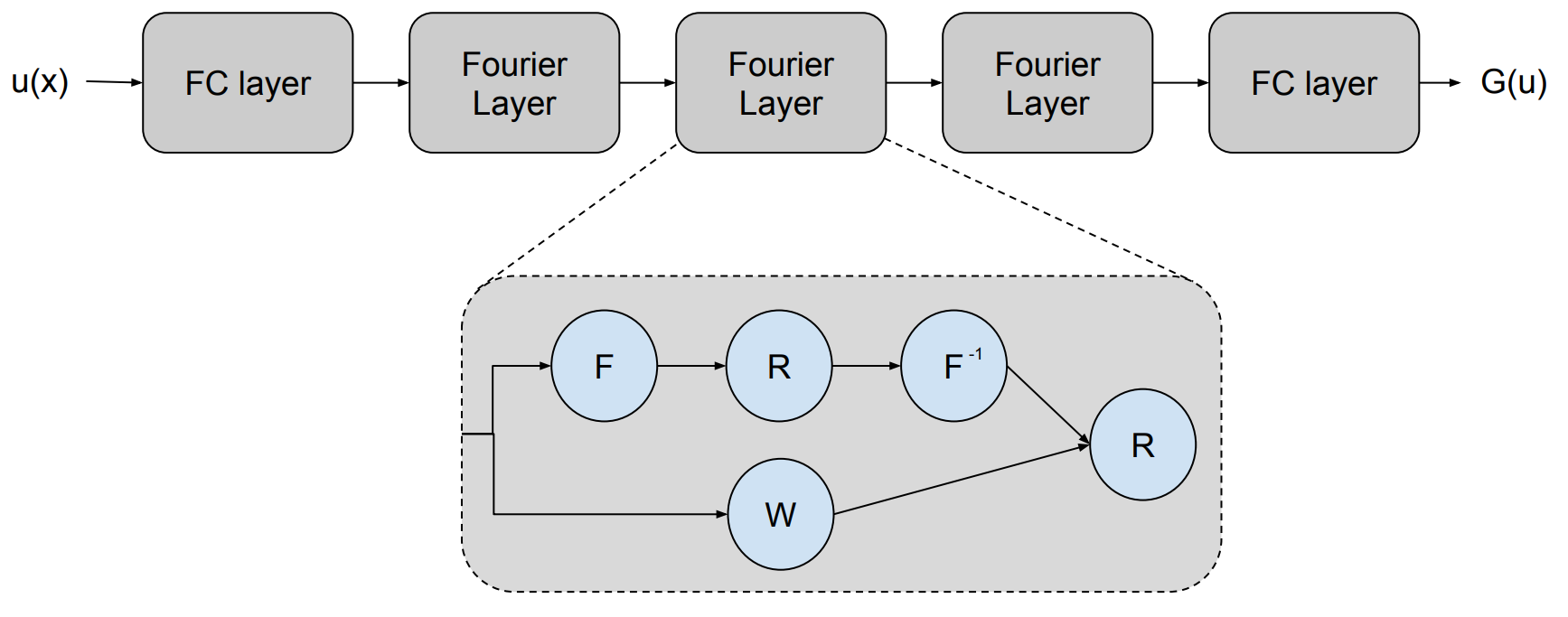}
        \caption{Fourier Neural Operator (FNO) \cite{li2020fourier}}
        \label{fig:linear_advection_compare}
    \end{subfigure}

    \caption{Benchmark methods are introduced to handle problems with discontinuities. HyperDeepONet and flexDeepONet incorporate additional connections that inform the input $u(x)$ into the trunk network. This modification reduces the limitations imposed by the final linear combination in the vanilla DeepONet.}
    \label{fig:comparision_models}
\end{figure}

Here, we briefly introduce the architectures used for comparison. These architectures are state-of-the-art and have been proposed in the literature to handle discontinuities (see Figure \ref{fig:comparision_models}). We then describe the experimental setup, where we intentionally use fewer trainable parameters for our Cut-DeepONet than in other methods. Hyperparameters are selected via iterative trial and error.

\begin{table}[h]
\centering
\small
\begin{tabular}{lcccc}
\toprule
\textbf{Method} & \textbf{Branch Net} & \textbf{Trunk Net} & \textbf{Additional Net} & \textbf{Params} \\
\midrule \midrule

\textbf{DeepONet} 
    & m-512-512-512-128
    & m-512-512-512-128 
    & 
    & 1440k \\ 

\textbf{flexDeepONet} 
    & m-256-256-256-129
    & m-256-256-256-128  
    & m-256-256-256-3  
    & 719k\\ 

\textbf{HyperDeepONet} 
    & m-150-150-150-4417 
    & 2-64-64-1 
    &
    & 787k\\ 

\textbf{FNO} 
    & 64x3
    & 4×(64, 12x12 modes)
    & 64-128-1
    & 4744k \\ 

\textbf{Cut-DeepONet} \textit{(our)} 
    & m-256-256-256-128
    & m-256-256-256-128 
    & m-64-64-64-dis\_n
    & 458k + 42k \\ 

\bottomrule
\end{tabular}
\caption{The number of trainable parameters for each method is reported for the \textbf{linear advection problem}. The FNO is an exception, as it uses a single flow network; its three blocks correspond to the lifting layer, Fourier layers, and projection layer.}
\label{tb:setting_table_1}
\end{table}

\begin{table}[h]
\centering
\small
\begin{tabular}{lcccc}
\toprule
\textbf{Method} & \textbf{Branch Net} & \textbf{Trunk Net} & \textbf{Additional Net} & \textbf{Params} \\
\midrule \midrule

\textbf{DeepONet} 
    & m-350-350-350-128
    & m-350-350-350-128 
    & 
    & 933k \\ 

\textbf{flexDeepONet} 
    & m-256-256-256-129
    & m-256-256-256-128  
    & m-256-256-256-3  
    & 975k\\ 

\textbf{HyperDeepONet} 
    & m-150-150-150-4417 
    & 2-64-64-1 
    &
    & 862k\\ 

\textbf{FNO} 
    & 64-64-64
    & 4×(64, 12x12 modes) 
    & 64-128-1
    & 4744k \\ 

\textbf{Cut-DeepONet} \textit{(our)} 
    & m-256-256-256-256
    & m-256-256-256-256 
    & m-40-40-40-dis\_n
    & 652k + 84k \\ 

\bottomrule
\end{tabular}
\caption{The number of trainable parameters for each method is reported for the \textbf{Inviscid Burgers' problem}. The FNO is an exception, as it uses a single flow network; its three blocks correspond to the lifting layer, Fourier layers, and projection layer.}
\label{tb:setting_table_2}
\end{table}

\begin{table}[h]
\centering
\small
\begin{tabular}{lcccc}
\toprule
\textbf{Method} & \textbf{Branch Net} & \textbf{Trunk Net} & \textbf{Additional Net} & \textbf{Params} \\
\midrule \midrule

\textbf{DeepONet} 
    & m-450-450-450-128
    & m-450-450-450-128 
    & 
    & 930k \\ 

\textbf{flexDeepONet} 
    & m-350-350-350-129
    & m-350-350-350-128  
    & m-350-350-350-3  
    & 832k\\ 

\textbf{HyperDeepONet} 
    & m-150-150-150-4417 
    & 2-64-64-1 
    &
    & 703k\\ 

\textbf{FNO} 
    & 80-80
    & 4x(80, 16 modes)
    & 80-128-1 
    & 446k \\ 

\textbf{Cut-DeepONet} \textit{(our)} 
    & m-256-256-256-128
    & m-256-256-256-128 
    & m-40-40-40-dis\_n
    & 331k + 9k \\ 

\bottomrule
\end{tabular}
\caption{The number of trainable parameters for each method is reported for the \textbf{Parsimonious Model}. The FNO is an exception, as it uses a single flow network; its three blocks correspond to the lifting layer, Fourier layers, and projection layer.}
\label{tb:setting_table_3}
\end{table}

\section{Lifting in Neural Field with Generalized Winding Number}
\label{sc:lifting_chang}
We briefly present the idea of Chang et al. \cite{Chang2025Lifting}, who introduced the use of the generalized winding number for the lifting method in neural fields, which was later extended to operator learning by us. In particular, let $\Omega \subset \mathbb{R}^2$ and $f : \Omega \to \mathbb{R},$ neural field with a discontinuity set $\Gamma \subset \Omega$. 
They introduce an augmented coordinate $z = H(x), x \in \Omega,$ where $H : \Omega \to \mathbb{R}$ is a function that is discontinuous across $\Gamma$ and constructed analytically. The new coordinate is defined as
\begin{equation}
\mathcal{T}(x) = (x, H(x)) \in \Omega \times \mathbb{R}.
\end{equation}
The original function is then represented as the restriction of the volumetric field:
\begin{equation}
f(x) = \tilde{f}_\theta(\mathcal{T}(x)) = \tilde{f}_\theta(x, H(x)), \quad x \in \Omega.
\end{equation}
Since $\tilde{f}_\theta$ is continuous in $\Omega \times \mathbb{R}$, it can be approximated by a network.

In their application of neural fields, discontinuities are caused by a curved, non-closed cut $\Gamma$. To encode the location of $\Gamma$, they employ the generalized winding number \cite{Jacobson2013Robust}, defining a lifting function $H(x)$. 
\begin{equation}
H(x) = \int_{\Gamma} \omega(x,y)\, d\sigma(y),
\end{equation}
where $\omega(x,y)$ measures the signed contribution of a boundary element $y \in \Gamma$ to the point $x$, and $d\sigma(y)$ denotes the surface measure on $\Gamma$. The generalized winding number $H(x)$ varies primarily in a neighborhood of the discontinuity set $\Gamma$, and remains approximately constant in regions far from $\Gamma$. If the cut is closed, $H(x)$ takes the value 0 in the exterior region and 1 in the interior region.

\section{Problem Equations}
\label{sc:problem_appendix}

\subsection{Linear Advection Equation}
This model describes the transport of a scalar quantity under a constant velocity field. It is governed by the linear advection equation:
\begin{equation} 
\frac{\partial u}{\partial t} + c \frac{\partial u}{\partial x} = 0,
\end{equation}
where $u(x,t)$ denotes the scalar field and $c$ is the constant advection speed, and $x \in [0,1]$, $t \in [0, 0.25]$ and $c = 1$. $u_0(x)= u(x,0)$ is a piecewise constant initial condition defined as:
\begin{equation}
u_0(x) =
\begin{cases}
h, & |x - s| \le \frac{w}{2}, \\
0, & \text{otherwise},
\end{cases}
\end{equation}
with pulse height $h \in [0.2, 0.8]$, width $w \in [0.2, 0.45]$, and midpoint $s \in [0.1, 0.2]$ determining the first discontinuity location.

\subsubsection*{Exact Solution}

The linearity of the equation admits an analytic solution given by pure translation of the initial condition:
\begin{equation}
u(x,t) = u_0(x - ct),
\end{equation}
which preserves the shape of the initial profile while shifting it at constant speed $c$.

\subsection{Inviscid Burgers' Equation}

This is a prototypical nonlinear hyperbolic conservation law widely used to study shock formation and nonlinear wave propagation. The inviscid Burgers' equation is given by:
\begin{equation}
\frac{\partial u}{\partial t} + u \frac{\partial u}{\partial x} = 0,
\end{equation}
where $u(x,t)$ denotes the scalar field (e.g., velocity) defined over space $x \in [-1, 1]$ and time $t \in [0,0.5]$.

This equation can also be written in conservation form:
\begin{equation}
\frac{\partial u}{\partial t} + \frac{\partial}{\partial x} \left( \frac{u^2}{2} \right) = 0,
\end{equation}

where $u_0(x) = u(x,0)$ is a piecewise constant initial condition with a single discontinuity.
\begin{equation}
u_0(x) =
\begin{cases}
u_L, & x < x_d, \\
u_R, & x \ge x_d,
\end{cases}
\end{equation}
with $x_d$ denoting the initial discontinuity location, and $u_L$, $u_R$ the left and right states, respectively.

\vspace{0.3em}
\noindent
In this work, we consider the specific case with $a > 0$:
\begin{equation}
u_L = a, \quad u_R = 0,
\end{equation}

\textbf{Exact Solution (Shock Case)} For discontinuous initial data with $u_L > u_R$, the solution develops a shock wave. The exact entropy solution is given by:
\begin{equation}
u(x,t) =
\begin{cases}
u_L, & x < x_s(t), \\
u_R, & x \ge x_s(t),
\end{cases}
\end{equation}
where the shock position evolves as:
\begin{equation}
x_s(t) = x_d + s\,t,
\end{equation}
and the shock speed $s$ is determined by the Rankine--Hugoniot condition:
\begin{equation}
s = \frac{f(u_L) - f(u_R)}{u_L - u_R} = \frac{1}{2}(u_L + u_R).
\end{equation}

For the present configuration ($u_R = 0$), this simplifies to:
\begin{equation}
s = \frac{u_L}{2}.
\end{equation}

\vspace{0.3em}
\noindent
Thus, the solution remains piecewise constant, with a discontinuity propagating at constant speed, capturing the nonlinear wave steepening and shock formation inherent to the inviscid Burgers dynamics.

\subsection{Parsimonious Model}
This is a reduced version of Hodgkin-Huxley model for action potential of rabbit ventricular cardiomyocytes \cite{HorgmoJaegeretal2023}. This model is structured around a system of three ODEs:

\begin{equation}
    \begin{aligned}
        C_m \frac{dv}{dt} &= -\left(I_{\mathrm{Na}} + I_{\mathrm{K}} + I_{\mathrm{stim}}\right), \\
        \frac{dm}{dt} &= \frac{m_{\infty} - m}{\tau_m}, \\
        \frac{dh}{dt} &= \frac{h_{\infty} - h}{\tau_h},
    \end{aligned}
\end{equation}

where $v$ is the membrane potential (mV), $C_m = 1~\mu\text{F}/\text{cm}^2$ is specific membrane capacitance. $I_{\mathrm{Na}}$ and $I_{\mathrm{K}}$ denote the current densities of the Na$^+$ and K$^+$ channels, and are defined by:

\begin{equation}
\begin{aligned}
I_{\mathrm{Na}} &= g_{\mathrm{Na}} m^3 h (v - v_{\mathrm{Na}}), \\
I_{\mathrm{K}}  &= g_{\mathrm{K}} e^{-b (v - v_{\mathrm{K}})} (v - v_{\mathrm{K}}),
\end{aligned}
\end{equation}

where $g_{\mathrm{Na}} = 11~\text{mS}/\text{cm}^2$, $g_{\mathrm{K}} = 0.3~\text{mS}/\text{cm}^2$ denote the maximum conductance densities of Na$^+$ and K$^+$ channels, with $v_{\mathrm{Na}} = 65~\text{mV}$, $v_{\mathrm{K}} = -83~\text{mV}$ representing potentials of the two channels. 
$m^3 h$ and $e^{-b(v - v_{\mathrm{K}})}$ denote the open probability of the Na$^+$ and the K$^+$ channels, respectively, with $b = 0.047~\text{mV}^{-1}$. And,

\begin{equation}
\begin{aligned}
m_\infty &= \frac{1}{1 + e^{(v - E_m)/k_m}}, 
& \tau_m &= 0.12~\text{ms}, \\
h_\infty &= \frac{1}{1 + e^{(v - E_h)/k_h}}, 
& \tau_h &= \frac{2 \tau_h^0 \delta_h e^{\delta_h (v - E_h)/k_h}}{1 + e^{(v - E_h)/k_h}}~\text{ms}.
\end{aligned}
\end{equation}

The other constants are defined as follows:
\begin{equation}
\begin{aligned}
E_m &= -41~\text{mV}, \quad k_m = -4.0~\text{mV}, \\
E_h &= -74.9~\text{mV}, \quad k_h = 4.4~\text{mV}, \\
\tau_h^0 &= 6.8~\text{ms}, \quad \delta_h = 0.8.
\end{aligned}
\end{equation}

The initial conditions are chosen as:
\begin{equation}
v(0) = -83~\text{mV}, \quad m(0) = 0, \quad h(0) = 0.9.
\end{equation}

The stimulus current density $I_{\mathrm{stim}}$ is used to trigger an action potential by increasing the membrane potential sufficiently to open the Na$^+$ channels. It is described as:
\begin{equation}
I_{\mathrm{stim}} =
\begin{cases}
a_{\mathrm{stim}}, & \text{if } t \ge t_{\mathrm{stim}} \text{ and } t \le t_{\mathrm{stim}} + d_{\mathrm{stim}}, \\
0, & \text{otherwise},
\end{cases}
\end{equation}

where $a_{\mathrm{stim}} (\mu\text{A}/\text{cm}^2)$, $t_{\mathrm{stim}} (\text{ms})$ and $d_{\mathrm{stim}} (\text{ms})$ are chosen to generate different stimulus scenarios. We generate 200 cases for training ($\approx 5\%$ of all cases), 50 for validation, and 50 for testing, with $T = 400~(\mathrm{ms})$ and $dt = 0.005~(\mathrm{ms})$.

\newpage
\section{Problem of Standard Metrics}
\label{sc:metric_problem}

\begin{figure}[h]
    \centering
    \includegraphics[width=1\textwidth]{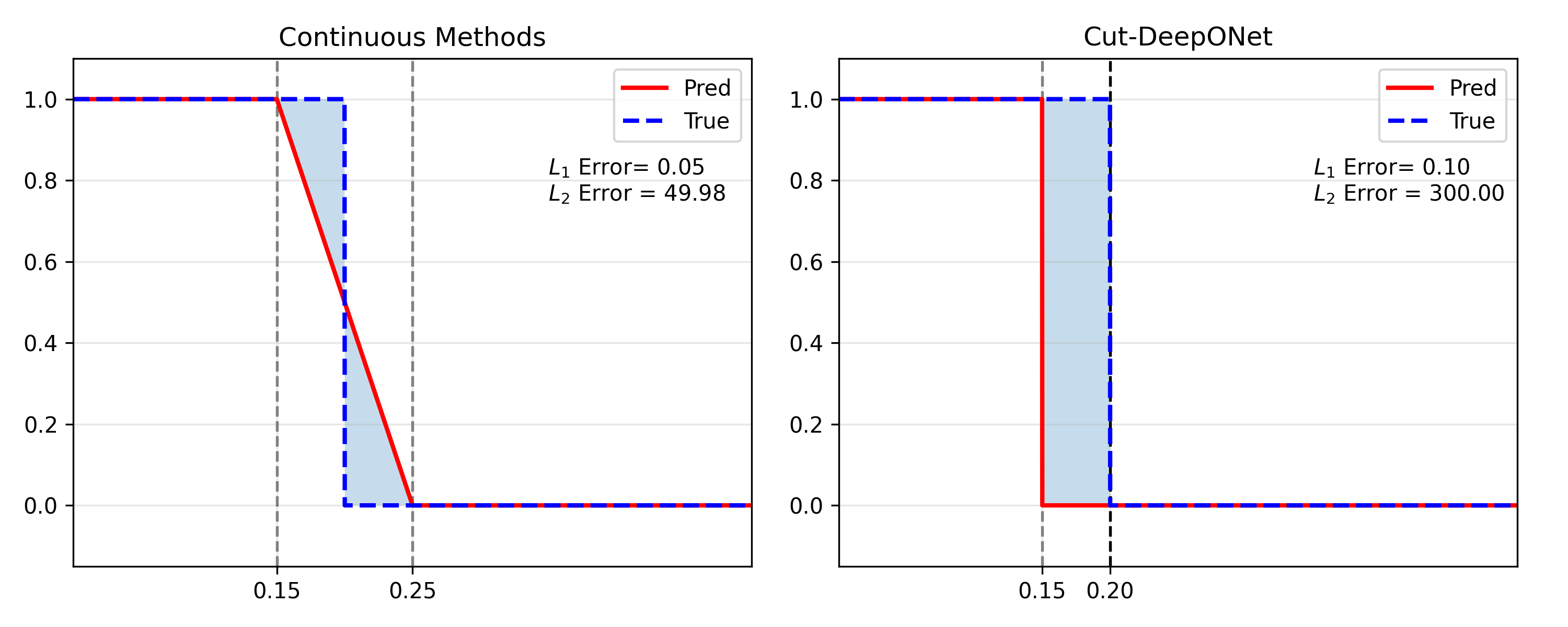}
    \caption{Cut-DeepONet generates discontinuous solutions, which have higher $L_1$ and $L_2$ errors than continuous methods, even though it is closer to the true solution. Nevertheless, Cut-DeepONet still outperforms the baselines in Table~\ref{tb:benchmark_table}; however, these metrics may underestimate its true effectiveness.}
    \label{fig:metric_problem}
\end{figure}

\newpage
\section{Data Preparation: Discontinuity Extraction}
\label{sc:data_preparing}
\subsection{Analytical Determination of Discontinuities in the Inviscid Burgers’ Equation}
\label{sc:analytically}
For scalar first-order nonlinear hyperbolic PDEs in conservation form, discontinuities can be found analytically by using the method of characteristics to identify where characteristics intersect and then applying the Rankine–Hugoniot condition to determine the shock speed. When an analytic expression for the shock location exists, the Cutting Net is unnecessary for unseen inputs.

For example, we consider the inviscid Burgers’ equation with the setup described above. For a general initial condition $u_0(x) = u(x,0)$, the characteristics satisfy \cite{Cameron2024Burgers}
\begin{equation}
\frac{dx}{dt} = u, \qquad \frac{du}{dt} = 0.
\end{equation}
Hence, $u$ remains constant along characteristics, yielding
\begin{equation}
x(t) = x_0 + u_0(x_0)\,t.
\end{equation}

Shock formation occurs when characteristics intersect, i.e., when the mapping $x = x_0 + u_0(x_0)t$ loses invertibility:
\begin{equation}
\frac{\partial x}{\partial x_0} = 1 + u_0'(x_0)\,t = 0.
\end{equation}
Thus, the first shock time with corresponding location is
\begin{align}
    &t_s = -\frac{1}{\min_{x_0} u_0'(x_0)}, \quad \text{provided } \min_{x_0} u_0'(x_0) < 0, \\
    &x_s(t_s) = x_0^\ast + u_0(x_0^\ast)\,t_s, \quad x_0^\ast = \arg\min_{x_0} u_0'(x_0).
\end{align}

After formation, the shock trajectory $x_s(t)$ is governed by the Rankine--Hugoniot condition:
\begin{equation}
\frac{dx_s}{dt} = s = \frac{f(u_L) - f(u_R)}{u_L - u_R}, \quad f(u) = \frac{u^2}{2},
\end{equation}
which simplifies to
\begin{equation}
\frac{dx_s}{dt} = \frac{1}{2}(u_L + u_R),
\end{equation}
where $u_L = u(x_s(t)^-, t)$ and $u_R = u(x_s(t)^+, t)$. The initial condition for the shock is given by $x_s(t_s)$.

\newpage
\subsection{Extraction of Discontinuities and Sharp Transitions from Data}
\label{sc:extract_discontinuity}
However, the analytical approach described above does not generalize to all PDEs, and is not applicable when no governing PDE is available. Therefore, we rely on data-driven methods to extract discontinuity locations for the lifting procedure and to enable the Cutting Net to generalize to unseen inputs.

\begin{figure}[h]
    \centering
    \begin{subfigure}{\textwidth}
        \centering
        \includegraphics[width=0.9\textwidth]{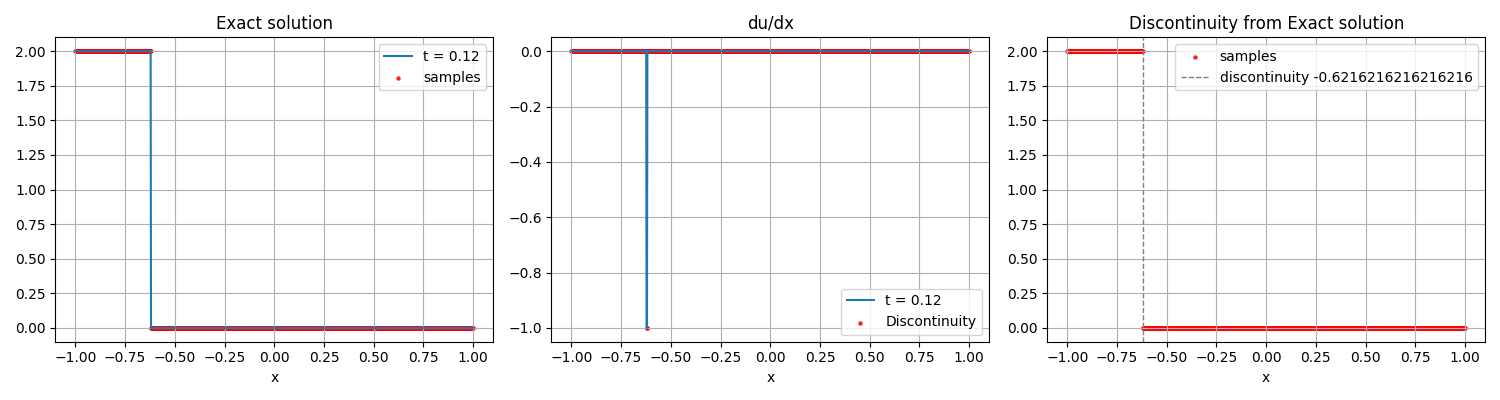}
        \caption{An exact solution is available for the inviscid Burgers’ equation.}
        \label{fig:data_discont}
    \end{subfigure}
    \begin{subfigure}{\textwidth}
        \centering
        \includegraphics[width=0.9\textwidth]{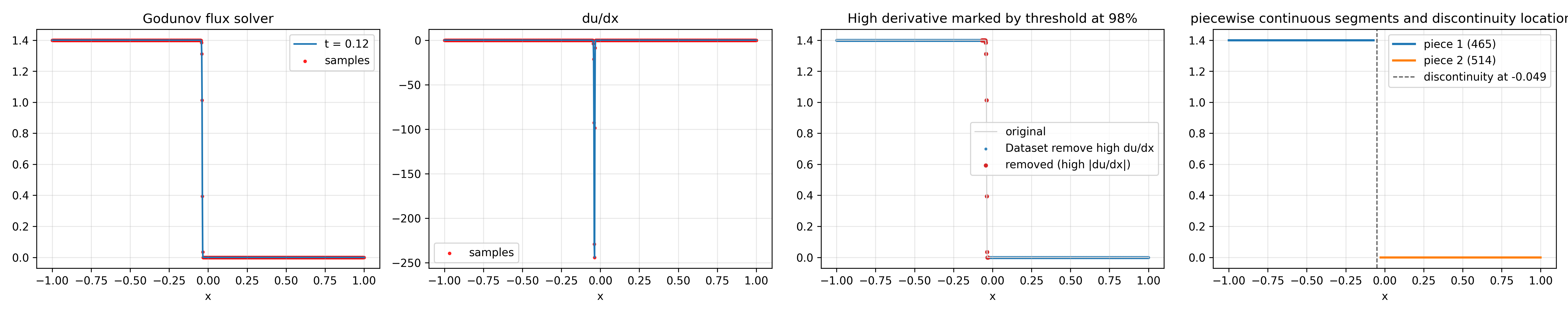}
        \caption{A numerical solution is obtained for the inviscid Burgers’ equation using the Godunov flux. The solution discretization from the numerical solver smooths out discontinuities, producing a continuous approximation. To mitigate its impact on operator learning, we filter out these regions from the training data.}
        \label{fig:data_discont}
    \end{subfigure}
    \begin{subfigure}{\textwidth}
        \centering
        \includegraphics[width=0.9\textwidth]{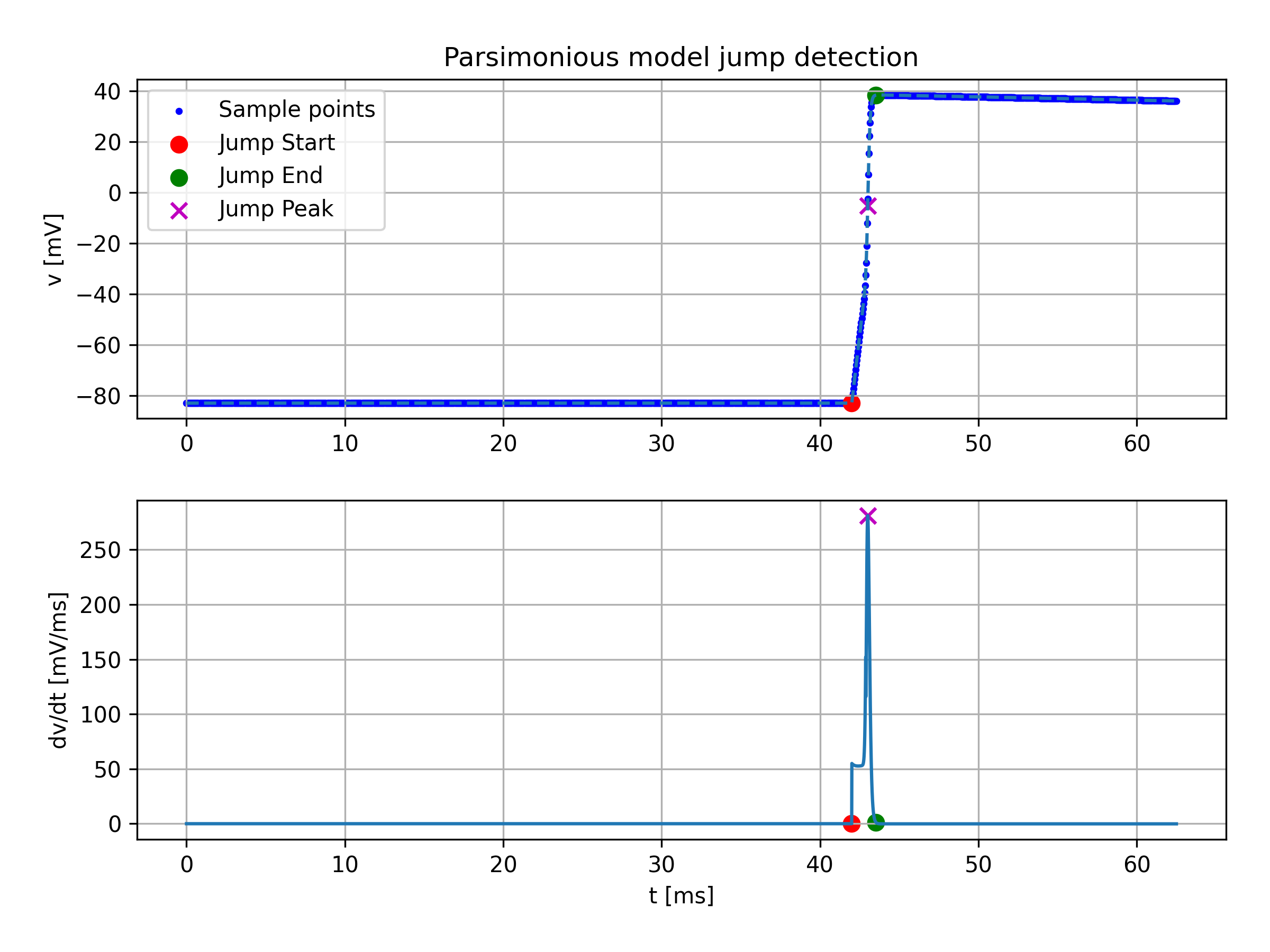}
        \caption{Sharp transition features are detected by identifying peaks and locating the nearest points where the derivative stabilizes.}
        \label{fig:sharp_extraction}
    \end{subfigure}

 
    \caption{Extraction of discontinuities and sharp transitions from data under different data availability scenarios.}
    \label{fig:example_burger}
\end{figure}

\newpage
\section{Convergence Tracking Comparison}

\begin{figure}[h]
    \centering
    \begin{subfigure}{\textwidth}
        \centering
        \includegraphics[width=0.9\textwidth]{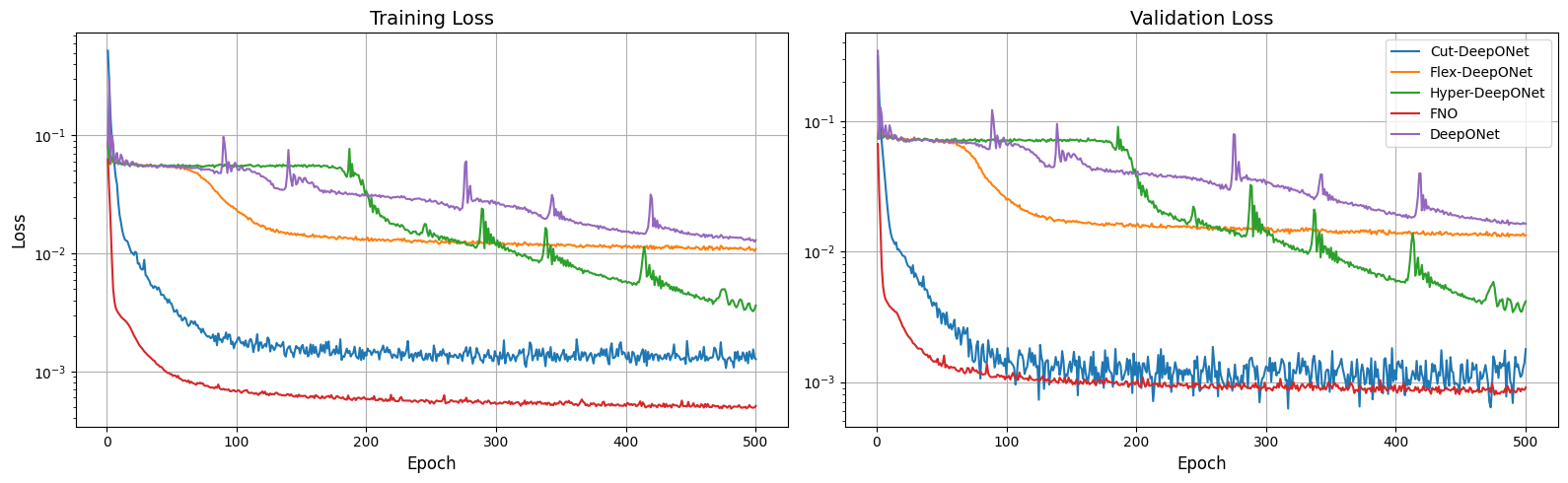}
        \caption{Convergence tracking for Linear Advection.}
        \label{fig:data_discont}
    \end{subfigure}
    \begin{subfigure}{\textwidth}
        \centering
        \includegraphics[width=0.9\textwidth]{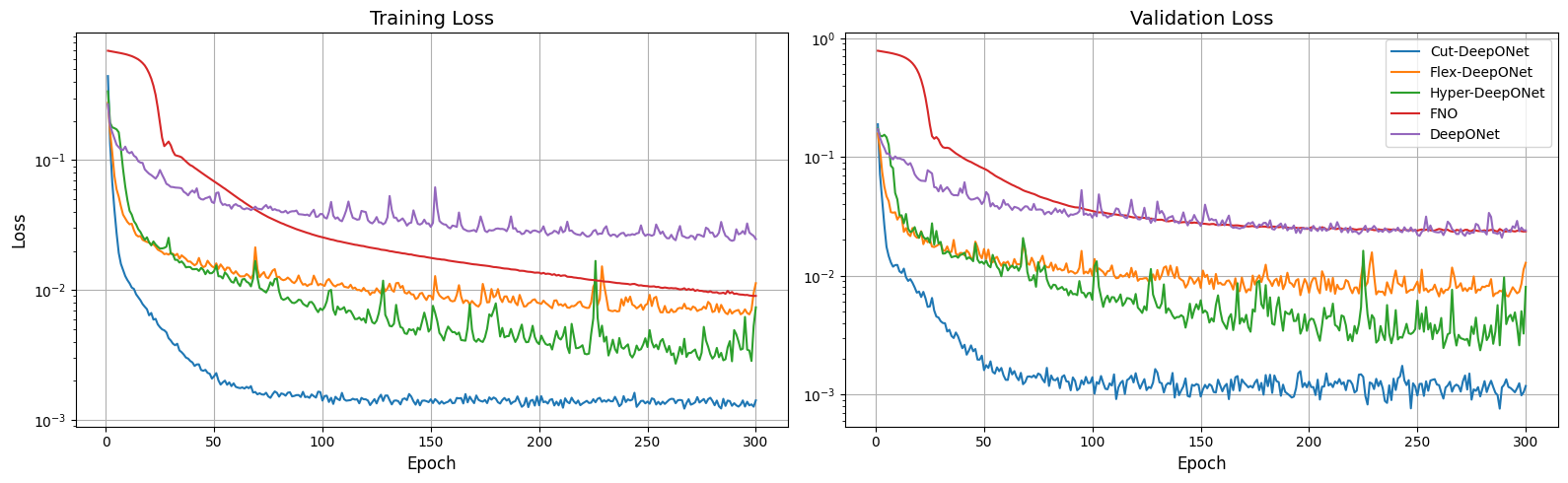}
        \caption{Convergence tracking for Inviscid Burgers' Equation.}
        \label{fig:data_discont}
    \end{subfigure}
    \begin{subfigure}{\textwidth}
        \centering
        \includegraphics[width=0.9\textwidth]{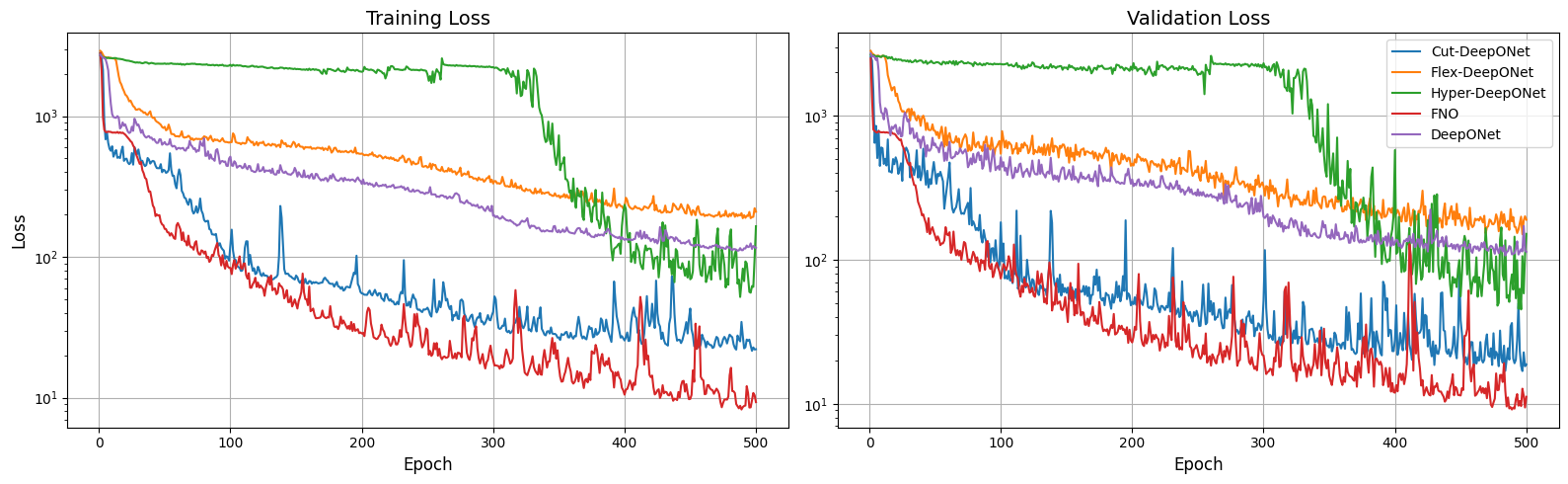}
        \caption{Convergence tracking for Parsimonious Model.}
        \label{fig:sharp_extraction}
    \end{subfigure}

 
    \caption{The convergence of different baseline methods is tracked across three tasks with a resolution of 2000. For Cut-DeepONet, the convergence corresponds to the second training stage. Cut-DeepONet converges faster than the other variants of DeepONet because it does not need to approximate discontinuities or sharp transitions. FNO uses a fixed grid mesh, making direct comparison with DeepONet difficult because testing on full and higher-resolution meshes is much more expensive. Therefore, the FNO curve is shown only to indicate convergence, not for direct comparison. For the inference test in Table~\ref{tb:benchmark_table}, we use the full resolution for all methods.}
    \label{fig:example_burger}
\end{figure}

\newpage
\section{Additional Experimental Visualization Results}

\begin{figure}[h]
    \centering
    \includegraphics[width=1\textwidth]{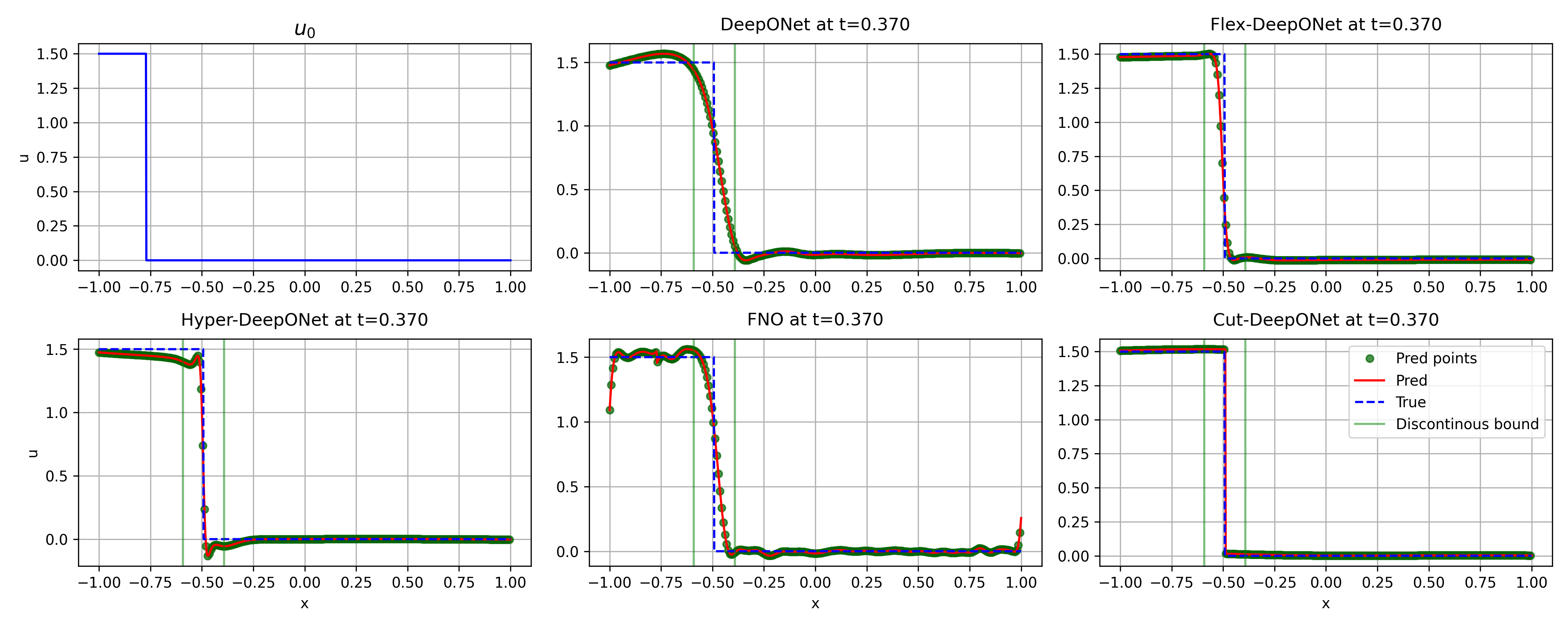}
    \caption{One example compares different methods on the Inviscid Burgers' Equation}
    \label{fig:linear_advection_result}
\end{figure}

\begin{figure}[h]
    \centering
    \includegraphics[width=0.95\textwidth]{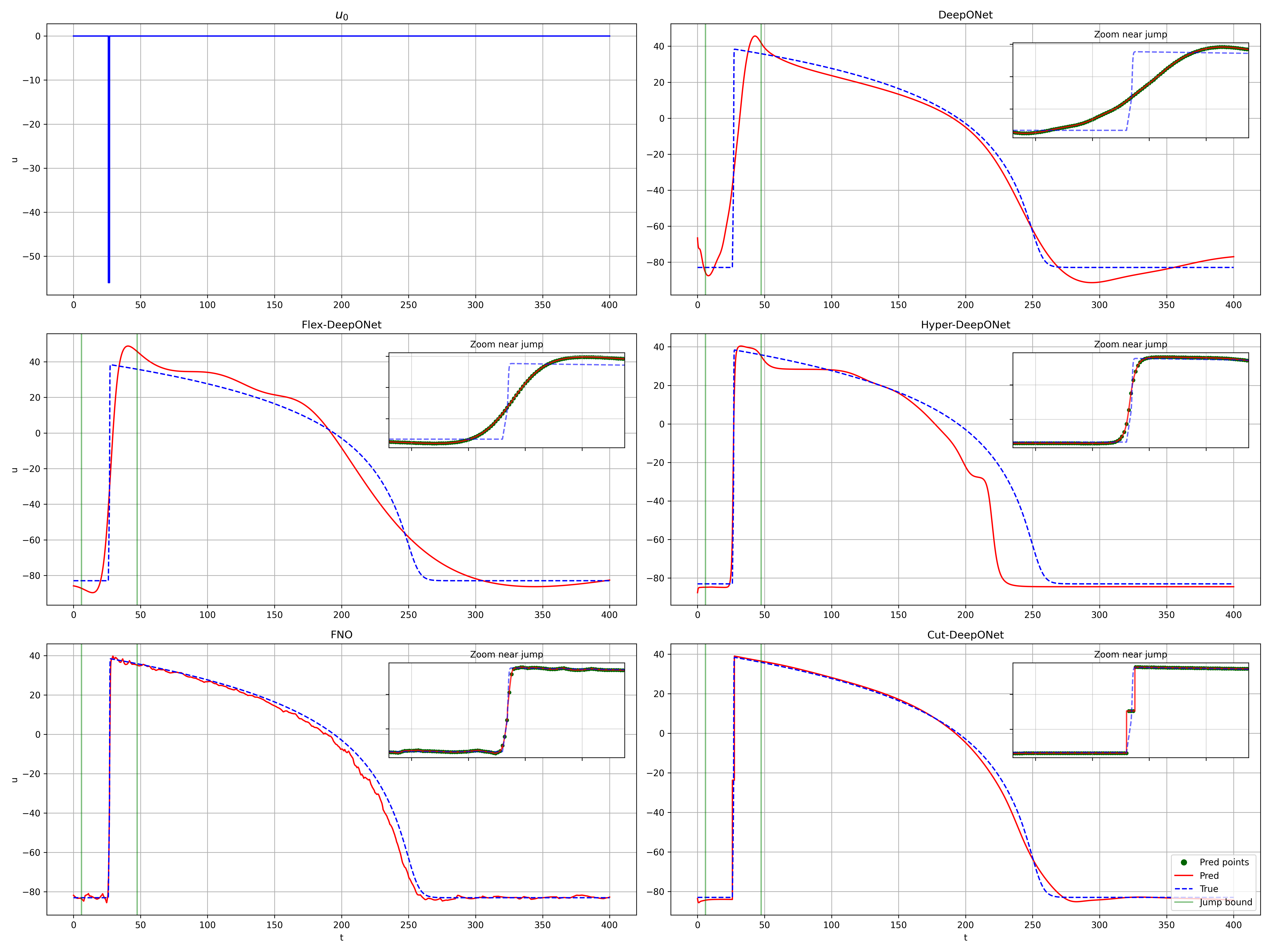}
    \caption{One example compares different methods on the parsimonious model. Regions near sharp transitions have a low probability of being sampled in the training data. Because the model is divided into separate pieces, the network cannot leverage data from neighboring regions to interpolate within a given piece. This limitation requires an additional sampling strategy to ensure that each piece has sufficient training data samples for learning.}
    \label{fig:pasimonious_result}
\end{figure}

\begin{figure}[h]
    \centering
    \begin{subfigure}{\textwidth}
        \centering
        \includegraphics[width=0.9\textwidth]{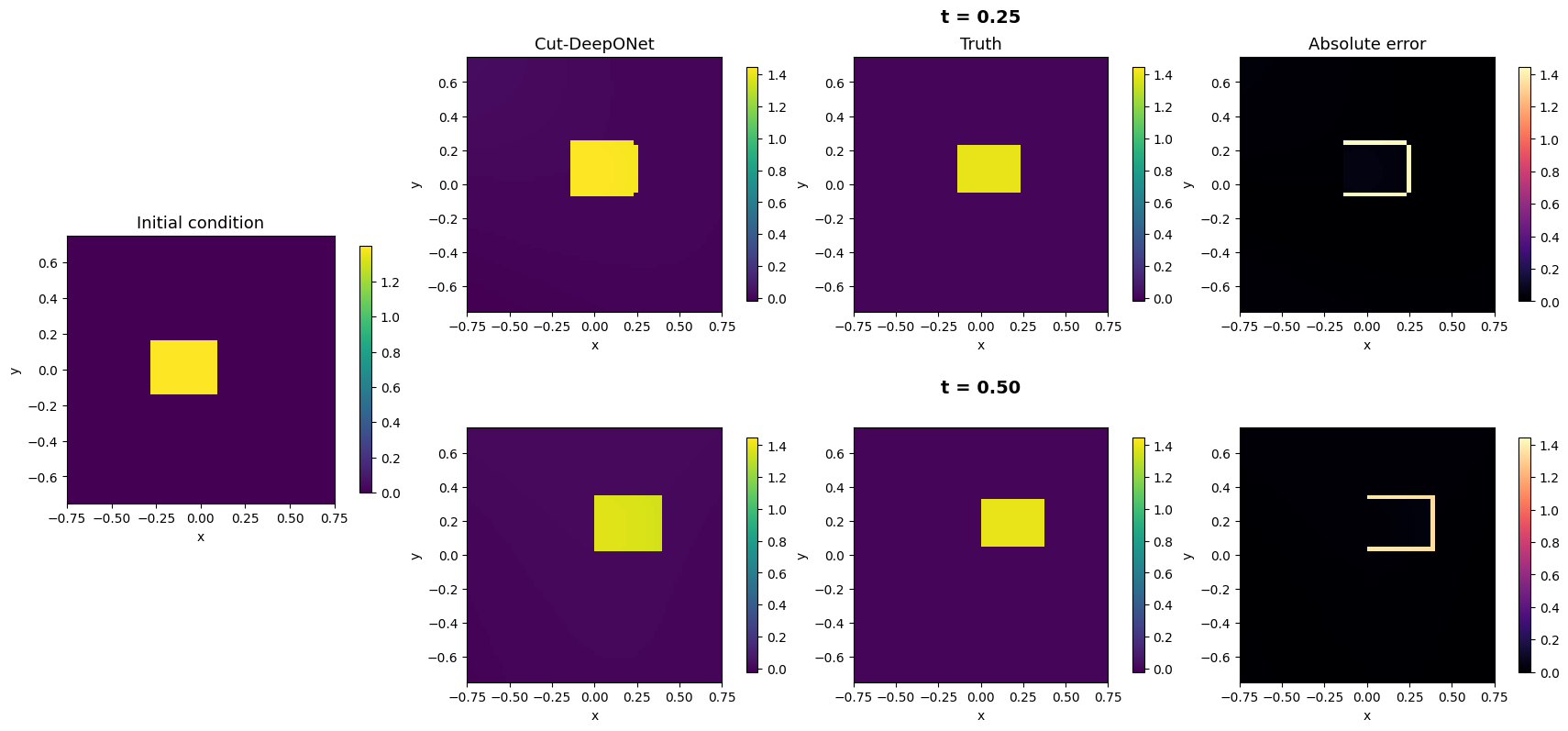}
        \caption{The error arises from the predicted cut locations, which is larger than the ground-truth cut. However, Cut-DeepONet reconstructs the correct smooth regions.}
        \label{fig:2d_linear_2}
    \end{subfigure}
    \begin{subfigure}{\textwidth}
        \centering
        \includegraphics[width=0.9\textwidth]{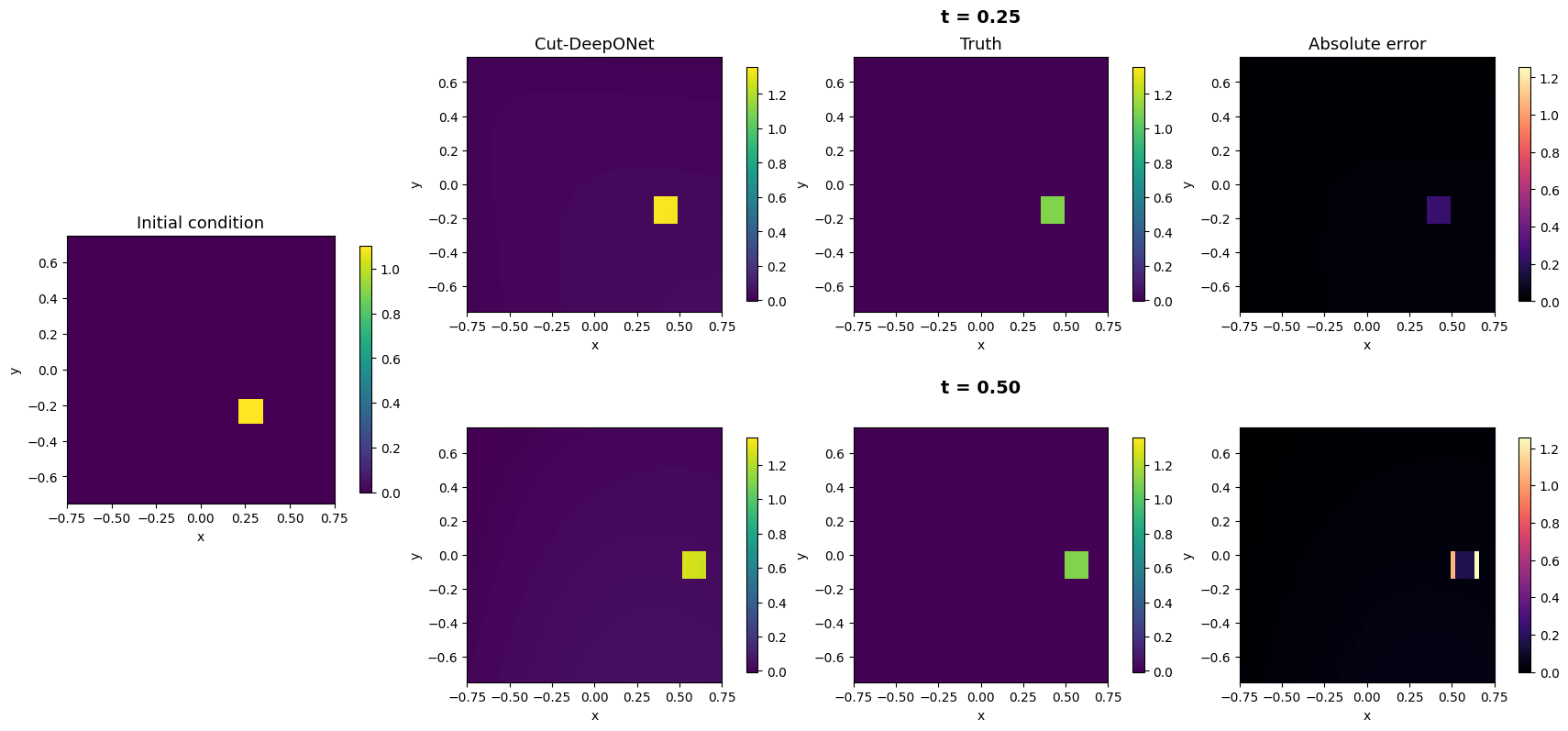}
        \caption{The error arises from both the incorrect magnitude predicted by Cut-DeepONet and inaccuracies in the predicted cut locations. Although the overall shape is still captured correctly, the error tends to increase when the lifted region becomes smaller. This may be related to the random sampling strategy, since only a few training samples are likely to fall inside these small regions, leading to insufficient representation during training, as also observed in the parsimonious case.}
        \label{fig:2d_linear_1}
    \end{subfigure}
 
    \caption{Cut-DeepONet extension for 2D linear advection. We consider 250 cases with varying initial rectangular sizes and locations, each represented on a \(64\times64\) grid. For training, approximately \(36\times36\) points are randomly sampled at each time step. For the Cutting Net, we use a U-Net architecture to predict a binary mask that identifies the cutting boundaries and segments the lifted regions. In Stage 2, the model is trained with an additional input dimension, \(H(x,y)\). The results demonstrate the advantages of the proposed method, illustrate its extension to higher dimensions, and show how the choice of cutting locations affects the final prediction.}
    \label{fig:2d_linear_advection}
\end{figure}

\begin{figure}[h]
    \centering
    \includegraphics[width=0.95\textwidth]{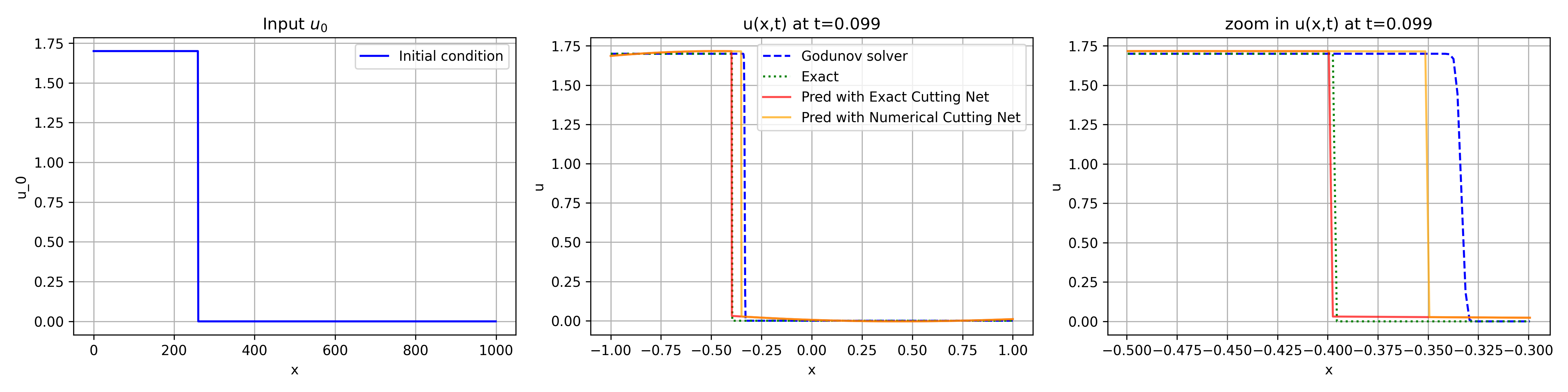}
    \caption{Low-resolution Godunov data are used for operator learning in Stage 2, while the Cutting Net in Stage 1 is trained using either exact-solution data or Godunov-solver data. The operator learning stage remains effective with low-resolution data, but the discontinuity prediction quality depends on the Cutting Net training data. The training setup is unchanged; only the dataset differs in this experiment.}
    \label{fig:cutting_net}
\end{figure}


\end{document}